\documentclass[lettersize,journal]{IEEEtran}
\usepackage[T1]{fontenc}
\newcommand{\kw}[1]{%
  {\fontfamily{lmtt}\selectfont
   \hyphenchar\font=`\-%
   \spaceskip=0.5em plus 0.16em minus 0.08em
   #1}}
\usepackage{amsmath,amsfonts}
\usepackage{algorithmic}
\usepackage[ruled,linesnumbered]{algorithm2e}
\usepackage{array}
\usepackage{bm}
\usepackage[caption=false,font=normalsize,labelfont=sf,textfont=sf]{subfig}
\usepackage{textcomp}
\usepackage{stfloats}
\usepackage{url}
\usepackage{verbatim}
\usepackage{graphicx}
\usepackage{cite}
\usepackage{tabularx,booktabs,makecell, multirow, multicol}

\usepackage{colortbl} 
\usepackage{xcolor}
\usepackage{amssymb}
\usepackage{bbding}
\usepackage{enumitem}

\definecolor{myblue}{RGB}{58, 104, 158}

\definecolor{myred}{RGB}{212, 45, 42}

\definecolor{LinkGrayBlue}{RGB}{64, 113, 139}

\definecolor{TableBlue}{RGB}{224, 236, 245}   
\definecolor{TableGray}{RGB}{239, 243, 246}   
\definecolor{TableStripe}{RGB}{248, 250, 252} 

\usepackage{hyperref} 

\hypersetup{
    colorlinks = true,
    pdfborder  = {0 0 0},
    linkcolor  = myred,        
    citecolor  = myblue,        
    urlcolor   = magenta   
}

\definecolor{TableHardwareHeader}{RGB}{151,170,188}
\definecolor{TableHighHeader}{RGB}{173,188,202}
\definecolor{TableLowHeader}{RGB}{193,204,214}
\definecolor{TableResultsHeader}{RGB}{211,219,226}
\definecolor{TableSubHeader}{RGB}{237,240,243}
\definecolor{TableStubHeader}{RGB}{212,219,226}
\providecolor{TableStripe}{RGB}{244,247,250}

\renewcommand{\epsilon}{\varepsilon}
\renewcommand{\phi}{\varphi}

\newcolumntype{L}[1]{>{\raggedright\arraybackslash}p{#1}}
\newcolumntype{C}[1]{>{\centering\arraybackslash}p{#1}}
\newcolumntype{R}[1]{>{\raggedleft\arraybackslash}p{#1}}

\makeatletter
\newcommand{\removelatexerror}{\let\@latex@error\@gobble}
\renewcommand{\maketag@@@}[1]{\hbox{\m@th\normalsize\normalfont#1}}%
\makeatother
\usepackage{makeidx}
\makeindex

\begin{document}
	\title{ Embodied Lower-limb Exoskeleton Assistance: Survey, Challenges, and Perspectives}
	\author{Xiao-Yin Liu,~\IEEEmembership{} Guotao Li*,~\IEEEmembership{} Robert Riener,~\IEEEmembership{} Zeng-Guang Hou*,~\IEEEmembership{}
        \thanks{ This work was supported by the National Natural Science Foundation of China under (Grant 62622321, Grant 62473365, Grant U22A2056, and Grant 62373013), the Noncommunicable Chronic Diseases-National Science and Technology Major Project (2024ZD0528000, 2024ZD0528002), the Beijing Natural Science Foundation under (Grant L232021, and L242101), and Talent Fund of the State Key Laboratory of Multimodal Artificial Intelligence Systems MAIS20260303. (*Corresponding authors: Guotao Li and Zeng-Guang Hou).}
        
		\thanks{Xiao-Yin Liu and Guotao Li are with the State Key Laboratory of Multimodal Artificial Intelligence Systems, Institute of Automation, Chinese Academy of Sciences, Beijing 100190, China, and also with the School of Artificial Intelligence, University of Chinese Academy of Sciences, Beijing 100049, China. (e-mail: liuxiaoyin2023@ia.ac.cn, guotao.li@ia.ac.cn).}

        \thanks{Robert Riener is with the Sensory-Motor Systems Lab, Department of Health Sciences and Technology, ETH Zurich, Zurich 8092, Switzerland. (e-mail: robert.riener@hest.ethz.ch).}

		\thanks{Zeng-Guang Hou is with the State Key Laboratory of Multimodal Artificial Intelligence Systems, Institute of Automation, Chinese Academy of Sciences, Beijing 100190, China, also with the School of Artificial Intelligence, University of Chinese Academy of Sciences, Beijing 100049, China, and also with CASIA-MUST Joint Laboratory of Intelligence Science and Technology, Institute of Systems Engineering, Macau University of Science and Technology, Macao, China. (e-mail: zengguang.hou@ia.ac.cn).}
	}
	\markboth{}%
	{Shell \MakeLowercase{\textit{et al.}}: A Sample Article Using IEEEtran.cls for IEEE Journals}
	\maketitle
	
	\begin{abstract}
    Over the past decade, artificial intelligence (AI) has enabled lower-limb exoskeletons to adapt to complex tasks and assist diverse users. Existing reviews, however, are organized around mechanical design, control strategies, or individual AI algorithms, and provide little systematic coverage of embodied lower-limb exoskeletons and their enabling technologies. 
    This review defines embodied lower-limb exoskeletons and organizes the literature around four dimensions: structural self-adaptation, task adaptation, user-specific adaptation, and control self-regulation. In each dimension, state-of-the-art and representative studies are classified and discussed in terms of the associated challenges, including lightweight and compliant design, data acquisition and fusion, optimization of complex policies, and robust, compliant control. The review then identifies the principal limitations of current research and outlines future directions along these dimensions.
    From an embodied perspective, this review sets out the closed-loop relationship among mechanical structure, sensing, and control as a framework for analyzing lower-limb exoskeleton performance and for research on human–robot interaction. The directions it identifies point toward lightweight, comfortable, and practical physical assistance.
    Project page: \href{https://xiaoyinliu0714.github.io/EmbExo/}{EmbExo}.    
    \end{abstract}

	\begin{IEEEkeywords}
	Embodied exoskeleton, Human-robot interaction, Task and individual adaptation, Efficient assistance
	\end{IEEEkeywords}

	\section{Introduction}
\IEEEPARstart{A}{dvances} in artificial intelligence (AI) have improved the ability of lower-limb exoskeletons to adapt to unstructured and dynamic environments, offering new opportunities to enhance walking ability in older adults and reduce the risk of joint injury \cite{molinaro2024task,shepherd2026roadmap}. However, exoskeletons operate within tightly coupled human–robot–environment systems, where assistance strategies must continually adapt to changes in the wearer’s movement, assistance needs, and environmental conditions \cite{luo2024experiment,liu2026exotraj}. These interdependencies make reliable and effective adaptive assistance challenging. To address these challenges, a growing body of research over the past five years has incorporated AI into exoskeleton systems to improve autonomous perception and decision-making, with the aim of improving the effectiveness of assistance \cite{molinaro2024estimating,van2025ai}.

\subsection{What is Embodied Lower-limb Exoskeleton Assistance?}
An embodied exoskeleton is a human-centered wearable intelligent agent capable of perception, decision-making, and adaptation to its operating context. In contrast to adaptive assistance, which adjusts the assistance strategy to the wearer, the task, or the environment, embodied assistance for lower-limb exoskeletons starts from the coupling among the human, the exoskeleton, and the environment. Mechanical structure, sensing, and control are brought into a single closed loop and act jointly: the exoskeleton integrates information about human–exoskeleton states, environmental conditions, and task requirements, and optimizes assistance torque profiles. The objective is efficient, comfortable movement assistance in dynamic environments~\cite{slade2022personalizing,molinaro2024estimating,scherpereel2025deep}. Because these elements share one loop, that objective requires mechanical structure, sensing, and control to be addressed jointly. These same three elements are the perspectives from which exoskeletons are studied and performance is analyzed.

Achieving the above assistance objective, however, presents four key challenges.
First, the mechanical structure must be lightweight, conform to human anatomy, and accommodate complex lower-limb joint kinematics to minimize additional interaction constraints and ensure wearing comfort \cite{xu2026bioinspired,yu2025autonomous}. Second, diverse assistance tasks and changing environments require assistance strategies that adapt to task demands and operating conditions \cite{molinaro2024task,luo2024experiment}. Third, differences in body morphology, physical capabilities, and gait characteristics necessitate assistance tailored to individual users \cite{lee2023user,scherpereel2025deep}. Finally, the bidirectional coupling between human and exoskeleton motion creates interdependencies between state perception and assistance control, requiring continuous optimization of assistance torques to balance assistance effectiveness and interaction comfort \cite{liu2026exotraj,huo2022impedance}. Accordingly, the core challenges for embodied lower-limb exoskeletons can be categorized into four areas: structural self-adaptation, task adaptation, user-specific adaptation, and control self-regulation.

To address these core challenges, existing research has pursued four complementary directions: joint self-alignment \cite{yu2025autonomous,sarkisian2020design} and compliant interaction \cite{li2026spinal,xu2026bioinspired} to improve structural self-adaptation; multimodal information fusion \cite{molinaro2024task,molinaro2024estimating} and musculoskeletal-informed policy learning \cite{luo2024experiment,luo2025robust} to support task adaptation; human-in-the-loop optimization \cite{slade2022personalizing,lee2023user,arens2025preference} and cross-subject transfer learning \cite{tan2026gaitdynamics,scherpereel2025deep} to enable user-specific adaptation; and control compensation \cite{sharifi2022trajectory,xing2024spatial} and human–exoskeleton control optimization \cite{walters2026optimal,liu2026exotraj} to improve assistance control. From a system architecture perspective, structural self-adaptation primarily concerns the physical design of the exoskeleton, task adaptation and user-specific adaptation primarily concern the high-level controller, and control self-regulation primarily concerns the low-level controller. These components serve as the “body,” “brain,” and “cerebellum” of the embodied system, respectively \cite{cangelosi2015embodied}. Functionally, an embodied lower-limb exoskeleton forms a closed-loop system that integrates structural design, perception and representation, and control optimization. These interdependent elements work together to support efficient and comfortable movement assistance in complex, dynamic environments.

\begin{figure*}
	\centering
	\includegraphics[width=1\textwidth]{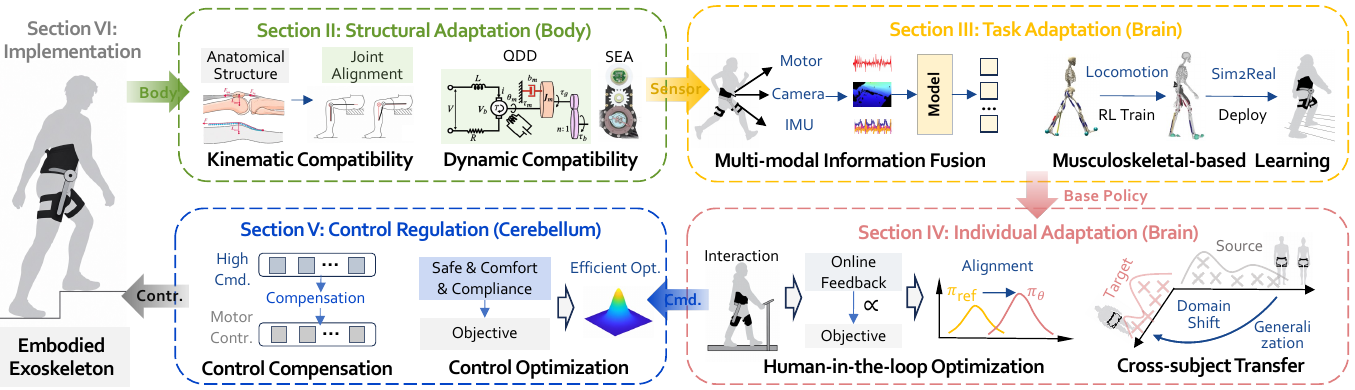}
	\caption{Organization and conceptual overview of the article. Section II addresses structural adaptation through kinematic and dynamic compatibility (physical embodiment). Sections III and IV constitute the high-level controller and focus on task and individual adaptation, respectively: task adaptation is achieved through multimodal information fusion or musculoskeletal-model-based training, whereas individual adaptation is enabled by human-in-the-loop optimization and cross-subject transfer. Section V presents the low-level controller, which achieves control self-regulation through control compensation and optimization. These three components jointly enable efficient assistance in complex environments (Section VI).} 
	\label{fig_intro}
\end{figure*}

\subsection{Summary of Previous Surveys}
Existing reviews of lower-limb exoskeletons have established a systematic knowledge base covering mechanical design, perception and control, and the evaluation of assistive performance. For mechanical design, early reviews surveyed device evolution and key technologies, including actuation, sensing, power supply, and control~\cite{dollar2008lower,young2016state}, and later reviews examined compliant mechanisms, attachment interfaces, bio-inspired structures, and human–robot interaction design~\cite{sanchez2019compliant,qiu2023systematic}. For perception and control, reviews of control and AI methods have addressed functional hierarchies, movement-intention recognition, user-specific parameter optimization, and reinforcement learning~\cite{baud2021review,li2021review,li2023human,10966209}, with increasing attention to gait planning that adapts to environmental conditions and to end-to-end, task-agnostic control~\cite{11360753,shepherd2026roadmap}. For evaluation, related reviews have broadened assistive outcomes to safety, embodiment, and physiological and clinical needs~\cite{wang2023evaluation,hybart2022embodiment,siviy2023opportunities}. Together, these reviews provide a foundation for classifying technologies and understanding individual methods.

However, this foundation has three limitations. First, because the reviews are organized by device type, functional module, or algorithm category, they offer limited analysis of how mechanical structure, sensing, algorithms, and control jointly support assistance through closed-loop interactions among the human, the exoskeleton, and the environment~\cite{shepherd2026roadmap}. Second, reviews that adopt an embodied perspective on perception, interaction, and cognition remain focused on the functions of individual technical modules, so perception, interaction, and cognition are not connected through that loop~\cite{xia2024shaping,10682963}. Third, the integration of AI into the loop has yet to be treated systematically, leaving the mechanisms for wearer comfort, changing task demands, individual differences, and the coupling between perception and control unclear. This review examines those mechanisms as four interrelated dimensions: structural self-adaptation, task adaptation, user-specific adaptation, and control self-regulation, together with their interdependencies and key challenges. In doing so, it provides a system-level perspective on personalized and comfortable assistance in complex, dynamic environments.

\subsection{Contributions and Paper Organization}
This review adopts an embodied intelligence perspective to examine three interconnected levels: physical embodiment, high-level decision-making, and low-level control, which functionally correspond to the system’s “body,” “brain,” and “cerebellum,” respectively. The embodied perspective also provides an analytical framework for this review of lower-limb exoskeletons.
Specifically, Section \ref{sec_mechanical} reviews the structural design of the physical embodiment. Sections \ref{sec_task} and \ref{sec_indi} examine task adaptation and user adaptation in high-level control, respectively, while Section \ref{Control} focuses on low-level control self-regulation. Building on these four parts, Section \ref{sec_imple} addresses practical system deployment, followed by an integrative discussion in Section \ref{sec_discus}.
Fig. \ref{fig_intro} illustrates the functional roles and interconnections of these components. The physical embodiment provides the foundation for sensing and control. High-level controllers based on AI algorithms adapt assistance policies to task requirements and individual user characteristics, providing desired trajectories or reference torques to the low-level controller. The low-level controller further optimizes the generation and regulation of assistive torques. Together, these interconnected levels support efficient assistance in complex real-world scenarios.

Each technical section reviews the literature around key challenges, comparing the characteristics and applicability of different approaches.
Section \ref{sec_mechanical} reviews human–exoskeleton kinematic compatibility \cite{xu2026bioinspired}, and lightweight design and compliant interaction \cite{qian2022toward}, addressing two challenges: accommodating complex joint kinematics and interindividual differences for efficient assistance (\textit{Challenge 1.1}), and reducing on-body mass and interaction impedance while preserving natural movement (\textit{Challenge 1.2}).
Section \ref{sec_task} examines task adaptation through multimodal information fusion \cite{molinaro2024estimating} and musculoskeletal policy learning \cite{luo2024experiment}. The former addresses low-cost multimodal data acquisition when torque is difficult to measure directly (\textit{Challenge 2.1}) and efficient fusion of high-dimensional multimodal features (\textit{Challenge 2.2}), covering torque \cite{molinaro2024task} and trajectory prediction \cite{liu2026exotraj}. The latter addresses training musculoskeletal policies in high-dimensional actuation spaces for complex terrain (\textit{Challenge 2.3}), and designing learning objectives and training frameworks for exoskeleton assistance that account for human–exoskeleton coupling (\textit{Challenge 2.4}), covering musculoskeletal locomotion \cite{Schumacher2023deprl} and exoskeleton assistance policies \cite{park2026learning}.

Section \ref{sec_indi} builds on these models and policies to examine user adaptation through human-in-the-loop optimization \cite{slade2022personalizing,lee2023user} and cross-subject transfer \cite{molinaro2024task,tan2026gaitdynamics}. The former addresses efficient algorithms for rapid optimization with limited interaction trials (\textit{Challenge 3.1}) and parameter optimization frameworks for complex network architectures that optimize policies on complex terrain (\textit{Challenge 3.2}), covering optimization based on physiological signals \cite{zhang2017human} and preferences \cite{liu2026pbbo}. The latter addresses training objectives that improve the generalization of personalized motion-intention prediction models (\textit{Challenge 3.3}), covering multi-subject data fusion \cite{molinaro2024task} and domain adaptation \cite{scherpereel2025deep}.
Section \ref{Control} reviews control compensation \cite{sharifi2022trajectory} and control optimization that support efficient assistance \cite{huo2022impedance}. It addresses accurate, real-time trajectory and torque tracking under uncertain human–exoskeleton systems (\textit{Challenge 4.1}), and control optimization under task completion, safety, and users’ voluntary movement (\textit{Challenge 4.2}).
These four aspects connect embodiment design, high-level adaptive decision-making, and low-level control self-regulation to support autonomous, adaptive assistance in complex, dynamic environments (Section \ref{sec_imple}). A comprehensive overview of the review framework is provided in Fig. \ref{fig_intro}.

\subsection{Literature Selection}
Studies on embodied lower-limb exoskeleton assistance published through October 2026 were identified in two stages.
In the first stage, IEEE Xplore and Google Scholar were queried along the four dimensions of embodied exoskeleton assistance. Each query combined \kw{lower-\allowbreak limb exoskeleton} or \kw{hip\slash knee\slash ankle exoskeleton} with one keyword set: 1)~Structural self-adaptation: \kw{kinematic\slash dynamic compatibility, lightweight design, compliant interaction}; 2)~Task adaptation: \kw{prediction, estimation, complex environment, musculoskeletal}; 3)~User-specific adaptation: \kw{human-\allowbreak in-\allowbreak the-\allowbreak loop optimization, preference, cross-subject}; 4)~Control self-regulation: \kw{compensation, tracking, optimization}.

In the second stage, the retrieved literature was restricted to representative studies published in major robotics journals and conferences. Within each dimension, priority was given to work from the most recent five years. Where that interval contained few relevant studies, the search window was extended further back and representative studies were retained. The resulting corpus concerns exoskeleton assistance that augments lower-limb strength during walking in healthy adults. Rehabilitation methods were included when their design or control principles extended to such walking assistance. Relevant arXiv preprints were added when the reported methods or experiments had not yet appeared in a journal or conference; these preprints are identified as such and are distinguished from peer-reviewed evidence.
	\section{Mechanical Design: Structural Adaptation}\label{sec_mechanical}

The mechanical design of a lower-limb exoskeleton governs kinematic compatibility, force transmission, and interaction comfort, while providing the foundation for sensing and control \cite{siviy2023opportunities,pruyn2026portable,zhang2026enhancing}. Achieving safe, compliant interaction is challenging because the knee combines rolling and sliding, whereas the hip has three rotational degrees of freedom \cite{yu2025autonomous}. Individual variations in joint geometry and anthropometry \cite{liu2024design} can make fixed-axis mechanisms produce pressure, shear, joint loading, and discomfort \cite{tricomi2024soft}. Meanwhile, actuators and power sources increase mass and energy demands, especially under sustained high-torque outdoor assistance \cite{archangeli2025lightweight,huang2022modeling}. Mass distribution affects effort and comfort, while excessive impedance restricts voluntary motion and disturbance recovery \cite{williams2025openexo,choi2025flexi}. These trade-offs motivate two key questions:
\begin{itemize}[leftmargin=*]
    \item \textit{Challenge 1.1}: How can exoskeletons accommodate complex, user-specific joint kinematics while transmitting assistance with minimal parasitic forces?
    \item \textit{Challenge 1.2}: How can lower-limb exoskeletons balance sufficient torque and power with low mass, distal inertia, and interaction impedance?
\end{itemize}

\begin{figure}
	\centering
	\includegraphics[width=0.5\textwidth]{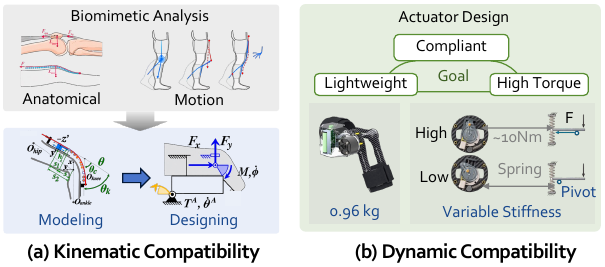}
	\caption{Schematic of two approaches to structural adaptation. (a) Kinematic compatibility: Inspired by human joint anatomy and kinematics \cite{xu2026bioinspired}, exoskeleton structures are designed to achieve compatibility with the wearer’s joints, for example through self-aligning mechanisms \cite{yu2025autonomous}. (b) Dynamic compatibility: Springs are incorporated into the joints to enhance interaction compliance, while lightweight designs improve interaction comfort \cite{li2026spinal}.} 
	\label{fig_sys}
\end{figure}

The two challenges correspond to i)~human–robot kinematic compatibility, including joint alignment and degree-of-freedom matching, and ii)~human–robot dynamic compatibility, involving the coordination of torque, mass, and impedance. Fig.~\ref{fig_sys} schematically illustrates the two categories of methods reviewed below.

\subsection{Human-Exoskeleton Kinematic Compatibility}
Lower-limb joints move as a coupled chain, and tibiofemoral rolling--sliding shifts the knee’s instantaneous center of rotation. Anatomical variation \cite{zanotto2015knee}, simplified joint models, and interface fit and placement \cite{bessler2022assessing} induce wearer--device relative motion and can raise internal joint load. Kinematic compatibility addresses \textit{Challenge 1.1} by accommodating this motion, transmitting the intended assistance, and limiting spurious loads \cite{zhang2024parallel}. Geometric-configuration compatibility supplies the static device--limb reference; motion-constraint compatibility is the set of constraints, reachable motions, and inter-joint coordination on that chain.

\subsubsection{\textbf{Geometric-Configuration Compatibility}}
It is the static device--limb correspondence that supplies the reference for kinematic compatibility. This correspondence is set by joint kinematic attributes, namely the assisted degrees of freedom (DoFs) \cite{li2024development} and the axis or instantaneous center of rotation (ICR) relative to the anatomical joint \cite{li2026bioinspired}, and by wearer-specific fit, including segment dimensions \cite{zanotto2015knee}, interface geometry \cite{lambranzi2025impact}, and soft-tissue compliance \cite{yandell2017physical}.

\textbullet\ \textit{Joint Kinematic Attribute Matching:}
It covers the assisted DoFs and the axis or ICR relative to the anatomical joint.
For the DoFs, hip flexion--extension is powered, while abduction--adduction remains passive and acts in series with the powered joint \cite{pruyn2026portable}. Releasing an unassisted rotation removes the parasitic constraint in that direction. Tracking migration of the knee center requires additional passive joints beyond these actuated axes, and each added joint increases mechanism complexity \cite{li2024development}. Knee mechanisms can also be synthesized for gait so that the device aligns with the flexion axis, while the optimization enforces the required mobility, self-alignment, and a desired force-transmission profile \cite{yu2025autonomous}.

For the axis, four-bar linkages approximate the rolling--sliding ICR \cite{kim2015design}. Their links can be dimensioned to a representative human ICR path, yielding one fixed geometry for that path. The same linkage can be fitted to a subject-specific curve reconstructed from medical images \cite{zeng2024design}; the fitted linkage matches one wearer and remains fixed after fabrication. Rolling gears place the ICR at the contact point and reproduce a prescribed roll--slide ratio \cite{yu2022bio}. A planetary stage approximates the rolling component, and a three-stage gear train compensates the residual sliding, reducing misalignment by about $70\%$ relative to a single-axis joint \cite{li2026bioinspired}. 
Gear trains reduce misalignment within a fixed roll–slide ratio, which remains subject-specific, and add transmission stages.

\textbullet\ \textit{Anthropometric Parameter Adaptation:} 
Three parameter groups are adapted.
Segment length and width are adjusted to the wearer. At equal knee-axis offset, a short robot thigh changed interaction force more than a long one \cite{zanotto2015knee}; thigh, shank, and torso lengths and widths were also adjusted on a hip–knee–ankle emulator \cite{bryan2021hip}. Length matching reduces force sensitivity, but the axis offset remains.
For the interface, semi-rigid pelvic and thigh shells are sized to each wearer \cite{pruyn2026portable}. Anchor position changes hip-flexor activation and metabolic cost, and no placement is optimal for every participant \cite{lambranzi2025impact}. Shells improve the static fit, whereas the effectiveness of anchor placement differs across people.
For soft tissue, interfaces absorbed about $55\%$ of the end-effector work during loading and returned most of it during unloading, delaying ankle augmentation \cite{yandell2017physical}. Compliance cushions the load but postpones assistance, so a rigid geometric match does not ensure timely torque delivery.

\subsubsection{\textbf{Motion-Constraint Compatibility}}
Motion-constraint compatibility requires that the coupled human--exoskeleton chain move without parasitic constraints. Once the geometric reference is fixed, it is addressed by closed-chain coordination \cite{cempini2012self,yu2025autonomous} and by matching reachable workspace to joint-center motion \cite{pruyn2026portable,asker2025optimum}.

\textbullet\ \textit{Closed-Chain Constraint Coordination:}
Exact-constraint design supplies a kinetostatic condition that delivers the intended torque while unloading misaligning degrees of freedom \cite{cempini2012self}. Knee mechanisms are synthesized to align with the flexion axis under simultaneous constraints on mobility, self-alignment, and force transmission \cite{yu2025autonomous}.
Passive serial chains decouple assisted rotation from misalignment: a passive prismatic--revolute--revolute chain in series with the knee axis \cite{sarkisian2020design}, and an actuated universal joint with one passive revolute and two passive prismatic joints \cite{li2024development}. These joints accommodate axis migration outside the assisted rotation.
Redundant revolutes and a linear guide passively accommodate the polycentric knee inside the joint \cite{xu2026bioinspired}. A parallel-elastic crossed four-bar reduces misalignment and unexpected interaction forces relative to a fixed-center commercial orthosis, and its springs supply partial gravity compensation \cite{zhang2024parallel}. The guide should match that path, and spring force can add unintended load.

\textbullet\ \textit{Workspace and Rhythm Matching:}
The two checks are workspace coverage and joint-center motion.
Passive abduction--adduction in series with powered hip flexion--extension admits frontal-plane displacement during sagittal assistance \cite{pruyn2026portable}.
Four-bar and rolling-gear joints follow a biological knee-center path \cite{kim2015design,yu2022bio}. Predicted thigh displacement of a fixed-center hinge grew with flexion angle. Hinges optimized for relative motion reduced that displacement even when their center differed from the average knee center \cite{asker2025optimum}.
Multiarticular textile paths \cite{asbeck2015biologically} and tendon-driven hip transmissions \cite{tricomi2024soft} follow limb motion without a rigid joint chain \cite{awad2017stroke}. Under hip-only assistance, the tested anchor configurations also changed knee and ankle kinematics \cite{lambranzi2025impact}.
Thus, the admitted frontal-plane displacement, the knee-center path, and the textile and tendon routes leave timing, relative displacement, and unassisted joint motion unresolved.

\subsubsection{\textbf{Summary}}
\textit{Challenge 1.1} is addressed through geometric-configuration compatibility and motion-constraint compatibility. The former sets the static device–limb reference via joint kinematic attribute matching (DoFs and ICR alignment) and anthropometric adaptation of segment dimensions, interface geometry, and soft-tissue compliance \cite{zanotto2015knee,yandell2017physical,yu2025autonomous,zeng2024design}. The latter then governs coupled motion without parasitic constraints through closed-chain coordination (exact-constraint mobility design and self-aligning redundancy) and workspace–rhythm matching of reachable ranges and physiological joint-center motion \cite{cempini2012self,pruyn2026portable,asker2025optimum,zhang2024parallel}.
Together, they allow the intended assistance to follow anatomical motion and to be delivered as effective joint torque.
However, separate geometric matching and constraint release leave subject-specific misalignment, soft-tissue delay, and added inertia unresolved. Future mechanisms should jointly synthesize subject-specific joint motion, self-aligning mobility, and interface dynamics on the coupled wearer--device chain.

\subsection{Human-Exoskeleton Dynamic Compatibility}

Lower-limb exoskeletons must balance sufficient torque and power with low added mass, distal inertia, and interaction impedance (\textit{Challenge 1.2})~\cite{veneman2006series,huang2022modeling,yan2025compact}. Two layers achieve this balance. The intrinsic layer keeps added inertia, weight, stiffness, and damping small~\cite{veneman2006series,yu2020quasi}. The coupled layer delivers the commanded torque and power under rapid loads.

\subsubsection{\textbf{Intrinsic Dynamic-Property Compatibility}}
Intrinsic dynamic-property compatibility is realized through mass distribution~\cite{yan2025compact} and mechanical impedance~\cite{yu2020quasi,qian2022toward}.

\textbullet\ \textit{Inertia and Mass Distribution:}
Three allocations reduce limb-borne mass and distal inertia while retaining torque and power: remote or proximal placement, on-joint integration, and a shared drive.
Remote placement moves the motors off the limb. In \cite{veneman2006series}, Bowden cables carried motor torque from off-board drives to a series-elastic joint. A remote knee actuator was developed in \cite{sulzer2009highly}, leaving $1.2$~kg on the limb. Proximal placement keeps the drive near the trunk. Soft hip shorts locate the motors on the rear waist belt, and tendons assist hip flexion \cite{tricomi2024soft}. A proximal motor drives a distal elastic module through a flexible shaft, whose torsional stiffness varies with configuration \cite{choi2025flexi}.

On-joint integration shortens the transmission. Quasi-direct drive (QDD) is applied at the hip \cite{yu2020quasi} and the knee \cite{huang2022modeling}. The actuator is enclosed in a carbon-fiber thigh link \cite{ishmael2022powered}, and peak torque is balanced with total mass and its placement, lowering bilateral mass \cite{yan2025compact}.
One actuator can serve two joint moments. One multiarticular soft path couples hip and ankle moments \cite{asbeck2015biologically,quinlivan2017magnitude,lee2017runningexosuit}. Knee extension and then ankle plantarflexion are assisted with one motor per leg through a continuum cable \cite{ma2021design}; a QDD motor with a gait clutch assists the knee and then the ankle \cite{wu2024novel}.
However, remote cables add friction and leave residual limb mass; proximal shafts add configuration-dependent stiffness; on-joint actuators keep mass on the limb; and a shared drive locks the assisted joint moments together.

\textbullet\ \textit{Mechanical Impedance:}
Interaction impedance is set by low-ratio actuation, series elasticity~\cite{veneman2006series,chen2024searehab}, and state-dependent stiffness or engagement.
Low-ratio QDD lowers reflected inertia and raises bandwidth \cite{yu2020quasi}. That bandwidth lets continuous stiffness-based torque replace discrete phase switching \cite{huang2022modeling}.
Series elasticity sets force bandwidth by spring stiffness, but cable friction leaves residual impedance \cite{veneman2006series}. Phase-specific knee flexion torque then holds impedance low outside that interval \cite{sulzer2009highly}. A nonlinear spring with zero-impedance control further reduces walking interaction torque \cite{qian2022toward}.
A parallel ankle spring first reduces motor current \cite{orekhov2022design}. A running clutch then returns stored negative work only while engaged \cite{sade2024design}. Extremum seeking next moves a lever arm as speed changes \cite{kumar2020extremum}, or two force-selected levels switch under simulated impairment \cite{aguirre2020lower}. Variable stiffness then sets isokinetic resistance and regenerates energy \cite{feng2025wearable}, and a backdrivable damping motor supplies portable resistance \cite{li2026spinal}.
However, continuous torque remains limited by winding temperature and cable friction, while the stiffness schedule is set by commanded force or a speed-dependent search.

\subsubsection{\textbf{Coupled-System Interaction Compatibility}}
Transmission design limits friction, torsion, and cross-axis load. Bowden cables and flexible shafts require compensation of cable friction or configuration-dependent torsion \cite{veneman2006series,sulzer2009highly,aguirre2020lower,choi2025flexi}. In \cite{archangeli2025lightweight}, parallel actuation keeps abduction independent of hip flexion; a serial chain redirects that torque into inversion and eversion at large flexion. A self-aligning interface is used against misalignment forces \cite{ishmael2022powered}. Shared drives leave the assisted joints coupled, whether one tension produces simultaneous hip and ankle moments \cite{asbeck2015biologically} or one motor produces knee and ankle moments in sequence \cite{ma2021design,wu2024novel}.
Load filtering attenuates impact and stores braking energy. A constant-force suspension that isolates impact on the back was developed in \cite{li2022mechanical}. A four-bar composite spring lowers torque and speed \cite{ishmael2022powered}. Clutched compliance is engaged during stance \cite{collins2015reducing} and during running braking \cite{sade2024design}. Natural frequencies relative to stride harmonics and soft-tissue resonance remain unidentified.

\begin{figure*}
	\centering
	\includegraphics[width=1\textwidth]{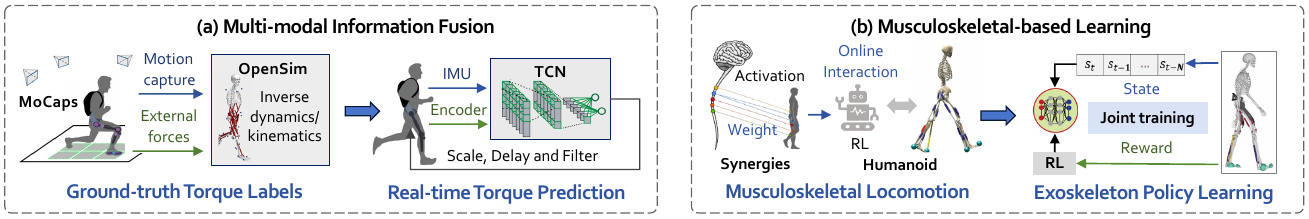}
	\caption{Schematic of two approaches to task-adaptive assistance. (a) Multimodal information fusion: Torque labels derived using motion capture are combined with multimodal sensor data from complex tasks to train neural networks for real-time torque prediction~\cite{molinaro2024task}. (b) Musculoskeletal-based learning: A musculoskeletal model is first trained to reproduce human-like walking, with bio-inspired strategies, such as muscle synergies, accelerating the training process~\cite{Schumacher2023deprl}. Reinforcement learning with a tailored reward function then jointly optimizes locomotion and exoskeleton assistance policies~\cite{luo2024experiment}.} 
	\label{fig_task}
\end{figure*}

\begin{table}[t]
    \normalsize
    \caption{Summary of representative mechanical designs for lower-limb exoskeletons.}
    \label{tab_mechanical_design}
    \centering
    \renewcommand{\arraystretch}{1.12}
    \setlength{\tabcolsep}{4pt}
    \arrayrulecolor{black}
    \resizebox{8.8cm}{!}{%
        \begin{tabular}{@{}l|ccc@{}}
            \toprule[1pt]
            \rowcolor{TableBlue}
            \multicolumn{4}{c}{\textbf{Human-Exoskeleton Kinematic Compatibility}} \\
            Related Works & Axis Alignment & Motion Fit & Load Path \\
            \midrule[0.3pt]
            \textcolor{magenta}{Yu et al. (2025)} \cite{yu2025autonomous}
            & Self-Alignment & DOF Synthesis & Force Opt. \\
            \rowcolor{TableStripe}
            \textcolor{magenta}{Xu et al. (2026)} \cite{xu2026bioinspired}
            & Redundant Joints & Linear Guide & Cable Drive \\
            \textcolor{magenta}{H. Li et al. (2026)} \cite{li2026bioinspired}
            & Gear Rolling & Slide Compensation & Gear Drive \\
            \rowcolor{TableStripe}
            \textcolor{magenta}{Yan et al. (2025)} \cite{yan2025compact}
            & Self-Alignment & Passive Slide & Thigh Brace \\
            \textcolor{magenta}{Lambranzi et al. (2025)} \cite{lambranzi2025impact}
            & -- & Soft Suit & Anchor Sites \\
            \midrule[0.7pt]
            \rowcolor{TableBlue}
            \multicolumn{4}{c}{\textbf{Lightweight Design and Compliant Interaction}} \\
            Related Works & Mass Saving & Drive Type & Compliance \\
            \midrule[0.3pt]
            \textcolor{magenta}{Huang et al. (2022)} \cite{huang2022modeling}
            & QDD Integration & Low Ratio & Low Inertia \\
            \rowcolor{TableStripe}
            \textcolor{magenta}{Ishmael et al. (2022)} \cite{ishmael2022powered}
            & Thigh Frame & 4-Bar Drive & Composite Spring \\
            \textcolor{magenta}{Y. Li et al. (2026)} \cite{li2026spinal}
            & Portable Mass & Damping Motor & Var. Stiffness \\
            \rowcolor{TableStripe}
            \textcolor{magenta}{Archangeli et al. (2025)} \cite{archangeli2025lightweight}
            & Light Frame & Parallel Drive & Indep. Planes \\
            \textcolor{magenta}{Choi et al. (2025)} \cite{choi2025flexi}
            & Proximal Motor & Flexible Shaft & Series Spring \\
            \bottomrule[1pt]
        \end{tabular}%
    }
\end{table}

\subsubsection{\textbf{Summary}}
\textit{Challenge 1.2} is addressed through intrinsic dynamic-property compatibility and coupled-system interaction compatibility. The former lowers limb-borne mass and distal inertia by remote or proximal placement~\cite{sulzer2009highly,choi2025flexi}, on-joint integration~\cite{ishmael2022powered,yan2025compact}, and a shared drive~\cite{asbeck2015biologically,wu2024novel}, and sets interaction impedance by low-ratio actuation~\cite{yu2020quasi,huang2022modeling}, series elasticity~\cite{veneman2006series,qian2022toward}, and state-dependent stiffness or engagement~\cite{aguirre2020lower,sade2024design}, while retaining torque and power. The latter then delivers that torque and power under rapid loads. Parallel actuation keeps abduction independent of hip flexion~\cite{archangeli2025lightweight}, and constant-force suspension and timed compliance attenuate impact and store braking energy~\cite{li2022mechanical,collins2015reducing,riemer2021extracting}.
Together, they balance sufficient torque and power with low mass, distal inertia, and interaction impedance.
Remote cables still add friction and residual limb mass~\cite{veneman2006series,sulzer2009highly}. Shared drives still lock the assisted joints, and winding temperature still limits continuous torque~\cite{ma2021design,yu2020quasi}. The natural frequencies of these coupled systems remain unidentified.

\subsection{Summary}
Lower-limb exoskeletons must accommodate anatomical variability while delivering sufficient assistance and minimizing mass and interaction impedance. 
Table~\ref{tab_mechanical_design} summarizes kinematic compatibility through alignment, motion accommodation, and load transfer, and dynamic compatibility through mass reduction, actuation, and compliance (impedance).
Existing studies have demonstrated improved kinematic compatibility through self-alignment, anatomical motion reconstruction, and adaptable interfaces \cite{sarkisian2020design,zeng2024design,lambranzi2025impact}. Proximal or shared actuation, elastic energy management, and low-impedance transmissions can further reduce mechanical burden and resistance to movement \cite{choi2025flexi,wu2024novel,orekhov2022design,huang2022modeling}. 
However, these benefits depend on the task and wearer: additional mobility increases mechanical complexity, interface deformation affects assistance transmission, and shared actuation constrains independent joint control \cite{zanotto2015knee,yandell2017physical,wu2024novel}. Future research should jointly optimize personalized kinematics, mass distribution, actuation, and interface design, supported by broader evaluations of delivered assistance, comfort, and energy efficiency across wearers and tasks to enable reliable everyday use.

	\section{High-level Controller: Task Adaptation} \label{sec_task}
Exoskeleton robots must assist with diverse tasks and terrains~\cite{lim2019delayed}. To this end, task-related observations $\mathbf{x}\in\mathbb{R}^{T_o\times N_o}$ are mapped to high-level control commands $\mathbf{y}\in\mathbb{R}^{T_p\times N_p}$, where $T_o$ and $T_p$ denote the observation and prediction horizons, respectively, and $N_o$ and $N_p$ denote the number of observation features and controlled joints. The mapping $\mathcal{F}$ is expressed as
\begin{equation}\label{eq_3_1}
    \mathbf{y}=\mathcal{F}(\mathbf{x}),
\end{equation}
where $\mathbf{y}$ typically comprises joint torques, joint angles, or related trajectory information, while $\mathbf{x}$ is obtained from inertial measurement units (IMUs), cameras, electromyography (EMG) sensors, joint encoders, and other relevant sensors.
Data-driven methods have been widely investigated to improve exoskeleton adaptability to complex tasks. However, acquiring multimodal data, particularly joint-torque data, remains a major challenge. This review examines task adaptation along two directions: 1) learning mappings from large-scale multimodal datasets~\cite{molinaro2024task,molinaro2024estimating}, and 2) training assistance policies in musculoskeletal-exoskeleton simulation environments~\cite{luo2023robust,luo2024experiment}. Fig.~\ref{fig_task} schematically illustrates these two approaches, which are discussed in the following sections.

\subsection{Multimodal Information Fusion}
Multimodal fusion methods use information from the coupled human-exoskeleton-environment system to predict joint torques~\cite{molinaro2024task}, joint angles~\cite{2026jointangle}, or motion trajectories~\cite{liu2026exotraj}. This information encompasses human motion and physiological states, exoskeleton mechanical signals, and environmental observations. Two key challenges remain:
\begin{itemize}[leftmargin=*]
    \item \textit{Challenge 2.1}: Because joint-torque data are difficult to measure directly, the first challenge centers on the low-cost acquisition of multimodal data.
    \item \textit{Challenge 2.2}: Given that multimodal data (e.g., imagery) are inherently high-dimensional, the second challenge lies in the efficient fusion of high-dimensional multimodal features.
\end{itemize}

Accurate prediction of torque or trajectory by the high-level controller does not amount to an optimal estimate of human movement intent. For trajectory tracking in particular, the torque profile used to follow the reference may diverge from the user’s movement pattern, so physical information about human motion must be incorporated. Whether these predictions yield efficient assistance depends on the exoskeleton structure and the low-level controller, and the predicted commands must be further optimized.


\begin{table*}
    \normalsize
    \caption{Related works and descriptions of multimodal datasets for lower-limb exoskeletons.}
    \label{tab_dataset}
    \centering
    \renewcommand{\arraystretch}{1}
    \setlength{\tabcolsep}{7pt}
    \arrayrulecolor{black}
    \resizebox{18.0cm}{!}{%
        \begin{tabular}{@{}l|cc|cl|c@{}}
            \toprule[1pt]
            \rowcolor{TableBlue}
            \textbf{Related Works} &
            \textbf{MoCaps} &
            \textbf{Sensors} &
            \textbf{Scenario} &
            \textbf{Tasks (Terrains)} &
            \textbf{Links}\\
            \midrule[0.4pt]

            \textcolor{magenta}{Camargo et al. (2021)} \cite{camargo2021comprehensive} &
            \Checkmark & IMU, sEMG, Force Plate & Indoor &
            Level ground, Stair and Ramp &
            \href{https://www.epic.gatech.edu/opensource-biomechanics-camargo-et-al/}{Data}\\

            \rowcolor{TableStripe}
            \textcolor{magenta}{Molinaro et al. (2024a)} \cite{molinaro2024estimating} &
            \Checkmark & IMU, Force Plate & Indoor &
            Level ground, Ramp and Stair &
            \href{https://datadryad.org/dataset/doi:10.5061/dryad.8kprr4xsv}{Data}\\

            \textcolor{magenta}{Molinaro et al. (2024b)} \cite{molinaro2024task} &
            \Checkmark & IMU, Force Plate & Indoor &
            Walk, Run, Stair climbing, Jump, Squat, Step up (28 activities)&
            \href{https://doi.org/10.24433/CO.7641031.v2}{Code} \\

            \rowcolor{TableStripe}
            \textcolor{magenta}{J. Li et al. (2026)} \cite{li2026k2muse} &
            \Checkmark & IMU, sEMG, Force Plate, AUS & Indoor &
            Level ground and Ramp walking &
            \href{https://k2muse.github.io/}{Data}\\

            \textcolor{magenta}{Luo et al. (2024)} \cite{luo2024experiment} &
            \XSolidBrush & IMU & Outdoor &
            Walk, Run and Stair climbing &\href{https://github.com/IntelligentRobotLearning/pseudocode_learning_in_simulation}{Code}\\

            \rowcolor{TableStripe}
            \textcolor{magenta}{Liu et al. (2026a)} \cite{liu2026exotraj} &
            \XSolidBrush & IMU, Encoder & Outdoor &
            Run, Rocky climb, Rugged terrain variable-speed walk, Jump and Squat &
            \href{https://xiaoyinliu0714.github.io/Home_ExoTraj/}{Page}\\
            \bottomrule[1pt]
        \end{tabular}%
    }
\end{table*}

\subsubsection{\textbf{Dataset Collection \& Description}} 
Most multimodal datasets for exoskeleton research are collected indoors because accurate joint-moment labels are difficult to obtain outside the laboratory. Early studies treated human joint moments computed with the OpenSim inverse-dynamics pipeline as locomotion ground truth~\cite{molinaro2024task,molinaro2024estimating}. Those estimates combine raw measurements from motion-capture systems (MoCaps) with six-axis force/torque transducers and plantar-pressure sensors. The equipment is specialized and labor-intensive, and it is costly and difficult to deploy in complex outdoor environments. Acquisition methods are therefore classified by whether they rely on motion capture.

Table~\ref{tab_dataset} summarizes representative recent multimodal datasets for human locomotion and lower-limb exoskeletons. Most exoskeleton datasets use MoCaps to estimate joint kinetics and kinematics, including torques and angles~\cite{molinaro2024estimating}, and representative sets~\cite{camargo2021comprehensive,molinaro2024estimating,molinaro2024task,li2026k2muse} mainly cover structured indoor terrains such as level ground, ramps, and stairs. Li et al.~\cite{li2026k2muse} additionally incorporated amplitude-mode ultrasound (AUS) and surface electromyography (sEMG), extending coverage to older adults and non-ideal conditions such as muscle fatigue. Collectively, these datasets support exoskeleton robotics, biomechanical analysis, and wearable sensing.

Motion-capture datasets therefore remain a source of indoor torque labels, while outdoor protocols trade those labels for deployability. Outdoor studies~\cite{liu2026exotraj,luo2024experiment} typically collect only IMU signals and motor-side measurements, such as joint angles and angular velocities, across grass, rocky surfaces, and mountain trails. This strategy substantially reduces acquisition cost and helps address \textit{Challenge 2.1}. Exoskeletons still require torque-level control, and deriving optimal torque commands from predicted joint angles or trajectories remains nontrivial. Reducing acquisition cost therefore shifts greater demands onto control algorithms. 

\subsubsection{\textbf{Torque Prediction}} 
Information fusion methods seek to learn the mapping $\mathcal{F}$ in Eq. \eqref{eq_3_1} from multimodal data $\mathbf{x}$ to torque or trajectory outputs $\mathbf{y}$. The dimensionality and complexity of these data make the mapping difficult to represent with simple mathematical functions. Existing methods therefore approximate $\mathcal{F}$ with a neural network $\mathcal{F}_{\omega}$ trained on the collected dataset $\mathcal{D}$, optimizing its parameters $\omega$ by minimizing prediction error:
\begin{equation}\label{eq_3_2}
    \arg\min_{\omega} \mathbb{E}_{\mathbf{x},\mathbf{y}\sim\mathcal{D}}\|\mathcal{F}_{\omega}(\mathbf{x})-\mathbf{y}\|_2,
\end{equation}
where $\mathbf{y}$ denotes the ground truth. When motion-capture-based torque labels are available, models such as temporal convolutional neural networks (TCNs) can be trained directly for torque prediction \cite{molinaro2024estimating,molinaro2024task}. To generate exoskeleton motor commands, these controllers scale the estimated biological moments by about $15\%$--$20\%$ and apply a delay, then use the scaled moment as the assistance torque. These methods predominantly make single-step predictions, for which simple temporal models can efficiently fuse multimodal features and accurately estimate torque. Their main challenge is obtaining ground-truth torque labels.

\textbullet \text{ } \textit{Torque (Label) Estimation:} MoCaps-derived torque data are, in essence, joint torque estimates. Estimation methods fall into two main categories: inverse dynamics (ID) methods \cite{2022id,2023id,molinaro2024task,molinaro2024estimating} and electromyography (EMG) methods \cite{2019emg,quesada2026emg,emg2021}.
ID methods estimate joint torque from limb kinematics and plantar pressure using musculoskeletal \cite{molinaro2024estimating,molinaro2024task} or mathematical models \cite{2023id,2022id}. Although widely regarded as the gold standard, they require MoCaps or inertial measurement units alongside costly plantar pressure plates, limiting feasible data collection settings.
EMG methods estimate torque from measured signals using Hill-type models \cite{2019emg}, nonlinear parametric regression \cite{quesada2026emg}, or neural networks \cite{emg2021}. However, interindividual variability requires subject-specific calibration, while limited EMG signal fidelity leaves estimation accuracy insufficient for practical exoskeleton assistance.

To reduce data acquisition costs (\textit{Challenge 2.1}) and improve torque estimation in dynamic environments, Liu et al.~\cite{liu2026torque} proposed an optimization-based method that directly estimates exoskeleton joint torques, rather than human joint torques, using only motor-side measurements of joint angles and angular velocities. Although this method enables low-cost estimation in complex environments, it relies on an exoskeleton dynamics model. Because high-fidelity modeling of human-exoskeleton coupling remains difficult, estimation accuracy may be limited.
Bird et al.~\cite{bird2026ultra} used ultra-wideband radar to estimate muscle forces noninvasively. Their method improves estimation accuracy under dynamic conditions and muscle fatigue, providing robust labels for exoskeleton assistance, but still requires relatively costly instrumentation. Further research on accurate and accessible torque estimation is therefore needed. The estimated torques can serve as labels $\mathbf{y}$ for training a prediction model $\mathcal{F}$ that provides real-time exoskeleton assistance.

\subsubsection{\textbf{Trajectory Prediction}} Trajectory prediction estimates future joint-angle trajectories $\mathbf{y}$ from past observations $\mathbf{x}$, after which motor torque commands $\mathbf{a}$ are regulated to track the predicted trajectories~\cite{11319164}. This approach reduces data acquisition costs but increases the complexity of low-level control~\cite{liu2026exotraj}. In most studies, users wear the exoskeleton while their joint motion is recorded through motor-side sensing.
Accurately predicting trajectories remains challenging because they are high-dimensional outputs. Existing methods can be classified as rule-based~\cite{8886526,tricomi2025leveraging,chen2024learning,dong2025,9737684} or end-to-end~\cite{liu2026exotraj,2025tf,2026vae}. Rule-based methods adopt compact trajectory parameterizations, such as dynamic movement primitives (DMPs)~\cite{chen2024learning,8967017}, probabilistic movement primitives (ProMPs)~\cite{dong2025}, Gaussian processes (GPs)~\cite{9737684}, or sinusoidal functions~\cite{tricomi2025leveraging}, such that only a small set of task-dependent parameters must be learned. In contrast, end-to-end methods use neural networks to predict high-dimensional trajectories directly from multimodal features~\cite{liu2026exotraj}. Efficiently fusing high-dimensional temporal or image data therefore remains a key challenge (\textit{Challenge 2.2}).

\textbullet \text{ } \textit{Rule-Based Methods:} Rule-based methods typically predict trajectories in two stages. Offline, recorded motion trajectories are compressed into task-specific parameters using DMPs~\cite{chen2024learning,8967017}, ProMPs~\cite{dong2025}, or GPs~\cite{9737684}. Online, multimodal sensor data are fused to identify the required assistance mode, while human-in-the-loop optimization refines the parameters used to generate the desired trajectory~\cite{chen2024learning,8886526}. To incorporate terrain information, Tricomi et al.~\cite{tricomi2025leveraging} classified images into uphill, downhill, stair-ascent, and stair-descent conditions, and then generated trajectories using terrain-specific amplitudes and frequencies of sinusoidal functions.
These methods are promising for rehabilitation because their parameters can be adapted to the patient's condition and rehabilitation stage. However, their reliance on parameters learned under specific static conditions limits performance in complex and dynamic environments.

\textbullet \text{ } \textit{End-to-End Methods:} End-to-end methods enable adaptive trajectory prediction in complex environments by extracting and fusing features from temporal human and exoskeleton motion data or terrain observations, and directly mapping them to high-dimensional trajectories~\cite{2026vae,liu2026exotraj}. Their accuracy depends on task complexity and interindividual variability. Existing approaches primarily employ long short-term memory (LSTM) networks~\cite{9395610}, convolutional neural networks (CNNs)~\cite{9446851}, Transformers~\cite{2026vae}, or CNN--LSTM architectures~\cite{2025tf,10.3389/fbioe.2024.1358022}. However, substantial gait variability, even within the same individual and task, limits the adaptability of conventional LSTMs and Transformers in complex, dynamic settings. To address this limitation, a fast flow-matching generative method has been introduced to improve prediction accuracy while preserving real-time performance~\cite{liu2026exotraj}. By learning the underlying data distribution, generative methods can further improve prediction generalization.

Terrain conditions also influence trajectory prediction within the coupled human-exoskeleton-environment system, yet efficiently fusing high-dimensional images with temporal data remains difficult (\textit{Challenge 2.2}). Most environment-aware methods~\cite{wu2023visual,wu2021,tricomi2025leveraging} have been evaluated only in fixed, structured settings, such as regular blocks and stairs. Image-conditioned trajectory prediction therefore requires further investigation in complex environments.
The highly dynamic human-robot system requires low-latency network inference~\cite{liu2026exotraj}, limiting the deployment of large, computationally intensive models on exoskeletons. In direct-drive systems, tight coupling between perception and control further complicates the conversion of predicted trajectories into exoskeleton torque commands.

\subsubsection{\textbf{Summary}} Information-fusion methods use neural networks to integrate multimodal features and predict joint torques~\cite{molinaro2024task,molinaro2024estimating}, trajectories~\cite{liu2026exotraj}, or trajectory parameters~\cite{tricomi2025leveraging,chen2024learning}. These methods require data spanning diverse tasks, and their network parameters are learned according to Eq.~\eqref{eq_3_2}.
Torque prediction can improve assistance in complex tasks but relies on costly equipment for data acquisition (\textit{Challenge 2.1}). Trajectory prediction reduces acquisition costs and is more suitable for complex outdoor environments, but introduces challenges in high-dimensional data fusion (\textit{Challenge 2.2}) and stable, robust control. Table~\ref{tab_infused} summarizes representative studies, classifying torque-prediction methods by the estimation and instrumentation used to obtain torque labels, and trajectory-prediction methods by their dimensionality-reduction and multimodal-fusion strategies. To convert predicted trajectories into torques, some methods employ impedance models to estimate interaction torques and use extended state observers (ESOs)~\cite{9737684} or disturbance observers (DOBs)~\cite{chen2024learning} for torque observation and final command generation. Overall, these methods fuse temporal motion data or environmental images to provide adaptive assistance across complex and dynamic tasks.

\subsection{Musculoskeletal-based Learning}
Musculoskeletal-based policy learning first trains a musculoskeletal model to reproduce human walking across varied terrain. An exoskeleton is then incorporated, and task-specific objectives are defined to train adaptive assistance policies \cite{luo2024experiment,luo2023robust}. These methods require no costly data acquisition equipment, and the resulting policies can be trained entirely in simulation. The musculoskeletal system is high-dimensional and overactuated, with human-like task execution achieved by modulating muscle excitation levels \cite{Schumacher2023deprl,he2024dynsyn}. Musculoskeletal-based exoskeleton assistance policy learning faces two challenges:
\begin{itemize}[leftmargin=*]
    \item \textit{Challenge 2.3}: Given the exceedingly high dimensionality of the actuation space, how can a musculoskeletal control policy be trained to generalize across complex terrains?
    \item \textit{Challenge 2.4}: Due to the inherent human–robot coupling, how can the objective function and training framework be designed for exoskeleton assistance policy learning?
\end{itemize}


\begin{table}
    \normalsize
    \caption{Summary of representative multimodal information fusion-based works for exoskeleton task adaptation.}
    \label{tab_infused}
    \centering
    \renewcommand{\arraystretch}{1}
    \setlength{\tabcolsep}{5pt}
    \arrayrulecolor{black}
    \resizebox{8.8cm}{!}{%
        \begin{tabular}{@{}l|ccc@{}}
            \toprule[1pt]
            \rowcolor{TableBlue}
            \multicolumn{4}{c}{\textbf{Torque Prediction}} \\
            {Related Works} & {Label Estimation} & {Device} & Predictor\\
            \midrule[0.3pt]
            \textcolor{magenta}{Molinaro et al. (2024a)} \cite{molinaro2024estimating} & ID-Based & MoCaps &TCN\\
            \rowcolor{TableStripe}
            \textcolor{magenta}{Molinaro et al. (2024b)} \cite{molinaro2024task} & ID-Based & MoCaps & TCN\\
            \textcolor{magenta}{Quesada et al. (2026)} \cite{quesada2026emg} & EMG-Based & EMG & -\\
            \rowcolor{TableStripe}
            \textcolor{magenta}{Gui et al. (2019)} \cite{2019emg} & EMG-Based & EMG & RBF\\
            \textcolor{magenta}{Liu et al. (2026c)} \cite{liu2026torque} & Optimization-Based & IMU, Encoder & TCN\\
            \midrule[0.7pt]
            \rowcolor{TableBlue}
            \multicolumn{4}{c}{\textbf{Trajectory Prediction}} \\
            {Related Works} & {Parameterization} & {Fusion Method} & Controller\\
            \midrule[0.3pt]
            \textcolor{magenta}{Chen et al. (2020)} \cite{9737684} & Gaussian Function & -- & ESO-Based\\
            \rowcolor{TableStripe}
            \textcolor{magenta}{Tricomi et al. (2025)} \cite{tricomi2025leveraging} & Sine Function & -- & PD\\
            \textcolor{magenta}{Y. Chen et al. (2024)} \cite{chen2024learning} & DMPs & MLP & DOB-Based\\
            \rowcolor{TableStripe}
            \textcolor{magenta}{Liu et al. (2026a)} \cite{liu2026exotraj} & -- & Transformer (FM) & MPC\\
            \textcolor{magenta}{Cui et al. (2026)} \cite{2026vae} & -- & Transformer& PD \\
            \bottomrule[1pt]
        \end{tabular}%
    }
\end{table}

\subsubsection{\textbf{Musculoskeletal Locomotion}}
Musculoskeletal models simulate human motion and enable analysis of its biomechanics, dynamics, and kinematics. Common simulation platforms include OpenSim \cite{seth2018opensim}, MuJoCo \cite{caggiano2022myosuite}, and Hyfydy \cite{Geijtenbeek2019}. OpenSim is widely regarded as a gold standard in biomechanics and provides high-fidelity musculoskeletal models, but its computational speed limits reinforcement learning (RL) training \cite{seth2018opensim}. MuJoCo supports faster RL training, although its musculoskeletal modeling accuracy is lower than that of OpenSim \cite{caggiano2022myosuite}. Hyfydy combines OpenSim-level modeling accuracy with speed comparable to MuJoCo and integrates with SCONE for policy learning \cite{Geijtenbeek2019}. OpenSim also provides a broad range of human musculoskeletal models, while \href{https://www.myosuite.org/}{MyoSuite} offers open-source code and tutorials for training policies on such models.

Training lower-limb exoskeleton assistance policies requires musculoskeletal models to perform human-like locomotion across varied terrain. Because the human lower limb contains more than $50$ muscles, the resulting action space makes policy training difficult (\textit{Challenge 2.3}). Simple control algorithms struggle to optimize so many actuators, motivating the use of RL \cite{Schumacher2023deprl,chiappa2023latent,he2024dynsyn,simos2025reinforcement}. Existing RL methods improve training efficiency and policy performance by reducing action-space dimensionality through muscle synergies \cite{he2024dynsyn,schumacher2025emergence,berg2024sar} and bio-inspired motor control \cite{Schumacher2023deprl,zuo2024self}, or by improving exploration \cite{chiappa2023latent,geibeta2024exciting}.
The reward function $R(\mathbf{s},\mathbf{a})$, which evaluates action $\mathbf{a}$ in state $\mathbf{s}$, is central to learning a human-like locomotion policy $\pi(\mathbf{a}|\mathbf{s})$. Studies have therefore designed different rewards to promote human-like walking. We next review RL-based musculoskeletal walking and methods for accelerating policy training.

\textbullet \text{ } \textit{Problem Formulation:} In the musculoskeletal context, the state $\mathbf{s}$ is provided by the simulation environment, which primarily comprises muscle excitation magnitudes, joint positions and velocities, pelvic position, and ground reaction forces. The action $\mathbf{a}$ is defined as the muscle excitation signal for each muscle \cite{luo2024experiment}.
The policy $\pi(\mathbf{a}|\mathbf{s})$ can be trained via the following formulation:
\begin{equation}\label{eq_3_3}
    \arg\max_{\pi}\text{ } \mathbb{E}_{\pi}\left[\sum\nolimits_{t=0}^T\gamma^t R(\mathbf{s}_t,\mathbf{a}_t)\right],
\end{equation}
where $\gamma$ is the discount factor, $T$ is the trajectory horizon, and $\mathbf{s}_t$ and $\mathbf{a}_t$ are the state and action at time $t$. To promote human-like walking, the reward typically combines imitation fidelity $R_{imit}$, postural stability $R_{stab}$, task completion $R_{task}$, energy expenditure $R_{exp}$, and movement smoothness $R_{smth}$ \cite{luo2024experiment,he2024dynsyn,berg2024sar,schumacher2025emergence}. Specifically, $R_{imit}$ measures the tracking error between model joint trajectories $\mathbf{q}$ and desired trajectories $\mathbf{q}_{des}$ recorded from human walking \cite{berg2024sar,he2024dynsyn,chiu2025learning}; $R_{stab}$ captures gait balance, typically through the center of pressure (CoP) $\mathbf{s}_{cop}$ derived from ground reaction forces \cite{luo2024experiment}; and $R_{task}$ assesses whether walking speed $v$ meets the commanded speed $v_{des}$ \cite{schumacher2025emergence, Schumacher2023deprl}. The remaining terms characterize aggregate muscle excitation magnitudes ($R_{exp}$) \cite{zuo2024self,clemente2024predictive} and temporal regularity of motion ($R_{smth}$) \cite{schumacher2025emergence}. Studies combine these terms according to their task specifications to train musculoskeletal walking policies.

Online RL optimizes Eq. \eqref{eq_3_3} to train complex walking policies. Most musculoskeletal RL frameworks use actor–critic methods, including Trust Region Policy Optimization (TRPO) \cite{geibeta2024exciting}, Proximal Policy Optimization (PPO) \cite{luo2024experiment,simos2025reinforcement}, and Soft Actor–Critic (SAC) \cite{he2024dynsyn,berg2024sar}. The actor learns the policy, while the critic learns a value function to guide policy updates.
Training is challenged by the high-dimensional muscle control space and the diversity of walking speeds and terrains \cite{he2024dynsyn,berg2024sar}. Applying PPO or SAC directly can lead to unstable training and fail to produce a viable walking policy. Researchers therefore incorporate action-space reduction \cite{he2024dynsyn,luo2024experiment} and exploration-enhancement techniques \cite{Schumacher2023deprl,geibeta2024exciting} to enable human-like walking. We discuss these strategies next.

\textbullet \text{ } \textit{Accelerating Training:}
\textit{Challenge 2.3} arises from both the high-dimensional control space and the diversity of walking tasks. Most studies \cite{luo2024experiment,he2024dynsyn, Schumacher2023deprl,geibeta2024exciting} address the former by training a musculoskeletal system to walk effectively under a specific task. Separate policies are therefore needed for different speeds and terrains. We first review methods that reduce action dimensionality or improve RL training efficiency.

To improve the effectiveness and efficiency of musculoskeletal policy training, existing methods either reduce control dimensionality through bio-inspired mechanisms \cite{he2024dynsyn,berg2024sar} or enhance RL exploration \cite{Schumacher2023deprl,chiappa2023latent,geibeta2024exciting,simos2025reinforcement}. Most dimensionality reduction methods draw on muscle synergies: fixed, predefined activation patterns of muscle groups established by the nervous system that simplify control and serve as building blocks for complex movements.
He et al. \cite{he2024dynsyn} used RL to train outputs represented through muscle synergies, simplifying motion generation and improving training efficiency. Berg et al. \cite{berg2024sar} combined direct RL outputs with muscle-excitation-mediated outputs in a dual-head action architecture, likewise improving training efficiency. However, both approaches require task-specific policies, limiting generalization across walking tasks.

Exploration-enhancing strategies either inject noise \cite{chiappa2023latent,simos2025reinforcement} or incorporate principles of biological locomotion \cite{Schumacher2023deprl}. Chiappa et al. \cite{chiappa2023latent} injected temporally correlated noise into the policy network’s latent state to improve performance. \mbox{Schumacher} et al. \cite{Schumacher2023deprl} integrated differential extrinsic plasticity (DEP), a self-organization method, into RL to accelerate musculoskeletal locomotion learning.
Although these approaches improve RL training efficiency in high-dimensional state spaces, few unified policies support diverse walking speeds and terrains. Existing unified policies for different speeds \cite{chiu2025learning} are limited to low-dimensional action spaces. Developing high-dimensional musculoskeletal locomotion policies for complex terrains therefore remains an open problem.

\subsubsection{\textbf{Exoskeleton Policy Learning}}
Building on musculoskeletal locomotion policies, exoskeleton assistance policies can be trained in simulation, reducing the costs of data collection and human-robot interaction~\cite{luo2024experiment,luo2025robust}. The process generally involves three steps: 1) importing the exoskeleton's physical and sensor models into the simulator, 2) training a torque-assistance policy, and 3) transferring the learned policy to physical hardware~\cite{chavarrias2025adaptive,luo2021reinforcement}. However, human-robot coupling makes it difficult to formulate objectives that quantify assistance performance and effectively guide policy training (\textit{Challenge 2.4}). We therefore review related work on assistance objectives and policy-training frameworks.

\textbullet \text{ } \textit{Optimization Objective Design:}
Training objectives for exoskeleton assistance should account for the human, robot, and task (\textit{Challenge 2.4}) \cite{choi2026musculoskeletal,luo2024experiment}. Similar to musculoskeletal objectives, they typically combine rewards for motion-task completion, reduced human physical burden, and stability, safety, and comfort \cite{Bonab2026Optimization, Wang2026Modeling}. Studies select and weight these terms to prioritize, for example, anthropomorphic assistance trajectories, reduced user effort, or greater comfort \cite{chavarrias2025adaptive}.
Existing works construct the optimization objective function via different combinations of these reward terms.

Motion-task rewards optimize exoskeleton torque to track reference joint angles, foot positions, and pelvic poses. An exponential transformation assigns higher rewards to smaller tracking errors \cite{luo2024experiment,luo2025robust,ratnakumar2026predicting}. This constitutes imitation learning aimed at human-like walking.
Human-burden rewards aim to reduce the user’s metabolic rate \cite{chavarrias2025adaptive}. Some studies constrain muscle activation, although simulated activation does not directly represent real metabolic rate \cite{luo2023robust}. Choi et al. \cite{choi2026musculoskeletal} instead estimated metabolic power from muscle activation, maintenance heat, heat associated with fiber shortening and lengthening, and mechanical contraction work, then penalized this estimate. Reducing muscular burden is beneficial only if the user can still complete the target motion.
Stability, safety, and comfort rewards limit aggressive policy outputs and enforce joint ranges of motion \cite{luo2024experiment,chavarrias2025adaptive}. Studies penalize first- or second-order differences between consecutive actions to promote smooth control, with larger penalties when joint torque exceeds thresholds. Future work should develop objective functions that better characterize assistance performance and guide more natural, efficient exoskeleton assistance.

\textbullet \text{ } \textit{Assistance Policy Learning:}  
Exoskeleton assistance policies are trained using two main paradigms. The first simultaneously trains the musculoskeletal gait policy and the exoskeleton control policy~\cite{luo2023robust,luo2024experiment,luo2025robust}. A unified objective jointly updates the interacting muscle-control and exoskeleton networks using proximal policy optimization (PPO) within a single simulation process~\cite{luo2023robust,luo2024experiment}. However, maintaining training stability remains difficult, particularly across complex tasks.
The second paradigm uses staged training: a musculoskeletal model is first trained to reproduce human-like gait, after which the exoskeleton policy is optimized separately or jointly with the gait policy~\cite{yuan2026smat,leem2026exo,park2026learning,choi2026musculoskeletal,yuan2026multi}. In one scheme, exoskeleton torque is initially set to zero while the muscle-driven gait policy is learned; torque actions are then introduced, and both policies are further optimized~\cite{park2026learning,choi2026musculoskeletal}. In another scheme, the gait policy is trained without the exoskeleton and then frozen while an initial low-torque assistance policy is learned; the two policies are subsequently fine-tuned jointly~\cite{yuan2026smat}. This combination of staged and joint training is promising for complex tasks.

To reduce the sim-to-real gap, domain randomization is commonly used during training to improve policy generalization~\cite{luo2024experiment}. Real-world assistance must also adapt to varying walking speeds and tasks. Although musculoskeletal approaches can assist at different speeds~\cite{li2026learning}, complex terrain remains challenging. High-dimensional, overactuated musculoskeletal models must first learn human-like locomotion on such terrain, requiring visual information to be incorporated into the simulation. An exoskeleton assistance policy can then be trained on the resulting gait policy.
Real-time inference is critical during physical deployment. Most existing studies fuse IMU and joint-angle signals for online prediction and employ model distillation to satisfy real-time computational requirements.

\begin{table}
    \normalsize
    \caption{Summary of representative musculoskeletal-based learning works for exoskeleton task adaptation.}
    \label{tab_mus}
    \centering
    \renewcommand{\arraystretch}{1}
    \setlength{\tabcolsep}{5pt}
    \arrayrulecolor{black}
    \resizebox{8.8cm}{!}{%
        \begin{tabular}{@{}l|ccc@{}}
            \toprule[1pt]
            \rowcolor{TableBlue}
            \multicolumn{4}{c}{\textbf{Musculoskeletal Locomotion}} \\
            {Related Works} & {Accelerating Type} & {Max. Action Dim.} & {Simulator}\\
            \midrule[0.3pt]

            \textcolor{magenta}{Chiappa et al. (2023)} \cite{chiappa2023latent} & Noise Injection & $39$ & MuJoCo\\
            \rowcolor{TableStripe}            
            \textcolor{magenta}{\mbox{Schumacher} et al. (2023)} \cite{Schumacher2023deprl} & Self-organization & $120$ & Hyfydy\\
            \textcolor{magenta}{Geiß et al. (2024)} \cite{geibeta2024exciting} & Exploration & $92$ & MuJoCo\\
            
            \rowcolor{TableStripe}             
            \textcolor{magenta}{Berg et al. (2024)} \cite{berg2024sar} & Muscle Synergy & $80$ & MuJoCo\\  
            \textcolor{magenta}{He et al. (2024)} \cite{he2024dynsyn} & Muscle Synergy & $700$ & MuJoCo\\
        
            \midrule[0.7pt]
            \rowcolor{TableBlue}
            \multicolumn{4}{c}{\textbf{Exoskeleton Policy Learning}} \\
            {Related Works} & {Training Type} & {Tasks} & Simulator\\
            \midrule[0.3pt]

            \textcolor{magenta}{Luo et al. (2024)} \cite{luo2024experiment} & Joint Train (PPO) & Walk, Run &  Hyfydy\\
            \rowcolor{TableStripe} 
            \textcolor{magenta}{Yuan et al. (2026)} \cite{yuan2026smat} & Staged Train (PPO) & Fixed Speed &  MuJoCo\\

            \textcolor{magenta}{ Leem et al. (2026)} \cite{leem2026exo} & Staged Train (PPO) & Fixed Speed &  MuJoCo\\
            \rowcolor{TableStripe}             
            \textcolor{magenta}{Park et al. (2026)} \cite{park2026learning} & Staged Train (SAC) & Varied Speed &  Hyfydy\\
        
            \textcolor{magenta}{B. Li et al. (2026)} \cite{li2026learning} & Staged Train (SAC) & Varied Speed &  MuJoCo\\
            \bottomrule[1pt]
        \end{tabular}%
    }
\end{table}

\subsubsection{\textbf{Summary}}

Simulation-based training reduces the costs of data collection and online human–robot interaction \cite{luo2024experiment}. However, the high-dimensional action space of musculoskeletal models and the complexity of terrain and assistance tasks make gait-policy training difficult (\textit{Challenge 2.3}) \cite{leem2026exo}. Bio-inspired mechanisms \cite{berg2024sar,he2024dynsyn} and improved action exploration \cite{chiappa2023latent,Schumacher2023deprl,geibeta2024exciting} have increased training efficiency, with promising results on single or multiple simple tasks. To better match assistance to users’ motion characteristics, studies have designed different objectives \cite{luo2023robust,yuan2026smat} and used joint \cite{luo2024experiment} or staged training \cite{park2026learning} of musculoskeletal and exoskeleton policies (\textit{Challenge 2.4}).
Table \ref{tab_mus} summarizes representative musculoskeletal-model-based approaches to lower-limb exoskeleton task adaptation. It compares musculoskeletal policy learning by acceleration method, action-space dimensionality, and simulation environment, and assistance policy learning by training scheme, algorithm, applicable task, and simulation environment. Extending these approaches from simple tasks to a broader range of complex tasks remains an open problem.

\subsection{Summary}
Diverse and complex environments require exoskeletons to adapt across assistance tasks. Existing research follows two main approaches: predicting user intent from multimodal data~\cite{molinaro2024estimating,molinaro2024task,liu2026exotraj} and learning musculoskeletal model--based assistance policies in simulation~\cite{luo2024experiment,luo2025robust,li2026learning}.
The first approach fuses information from the human-exoskeleton-environment system to predict joint torques~\cite{molinaro2024task} or motion trajectories~\cite{liu2026exotraj}. The resulting models are comparatively easy to train and deploy across complex tasks, but the approach requires either labeled torque data or effective fusion of high-dimensional features for trajectory prediction~\cite{tricomi2025leveraging}. The second approach simulates human locomotion and trains assistance policies using predefined objectives~\cite{park2026learning}. It reduces data acquisition and physical human-exoskeleton interaction costs, but training high-dimensional musculoskeletal control policies for complex tasks remains difficult~\cite{yuan2026smat}. Both approaches show promise for practical assistance, although reducing data acquisition and network-training costs remains an important research goal.
    \section{High-level Controller: Individual Adaptation} \label{sec_indi}
Interindividual differences in the neuromusculoskeletal system lead to distinct motor preferences, which can be inferred indirectly from joint torques and gait trajectories \cite{kim2019reducing,zhang2017human}. These differences can be viewed as variations in motor distributions. Individual adaptation seeks subject-specific parameters $\omega$ for the observation-to-command mapping $\mathcal{F}_{\omega}:\mathbf{x}\mapsto\mathbf{y}$:
\begin{equation}
    \omega^\star=\arg\max_{\omega}\mathcal{J}(\mathbf{x},\mathcal{F}_{\omega}),
\end{equation}
where $\mathcal{J}$ evaluates the command $\mathbf{y}$ generated from state $\mathbf{x}$ for a given user. Performance can be measured by how closely the distribution of generated actions matches the user’s motor distribution, by physiological measures of assistance such as metabolic rate \cite{2018science}, or by user preference feedback \cite{lee2023user}.

\begin{figure*}
	\centering
	\includegraphics[width=1\textwidth]{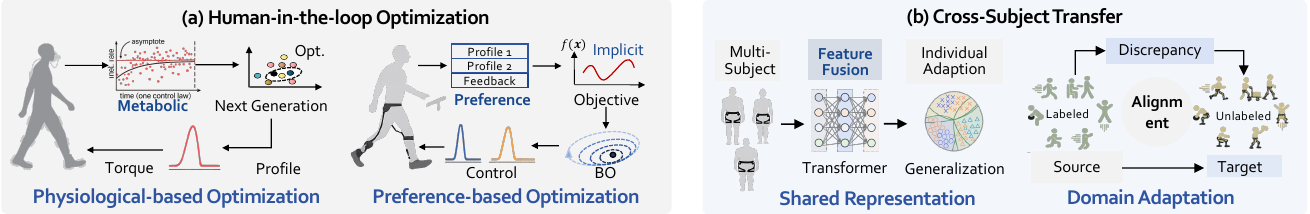}
	\caption{Schematic of two approaches to personalized assistance. (a) Human-in-the-loop optimization: User-specific parameters are iteratively updated to minimize metabolic rate~\cite{zhang2017human} or maximize a latent utility inferred from user preferences~\cite{lee2023user}, with feedback collected through online human--robot interaction. (b) Cross-subject transfer: Neural networks learn shared representations from multimodal data across subjects to improve generalization~\cite{molinaro2024task}, or align feature distributions between source and target participants to improve prediction accuracy for the target participant~\cite{scherpereel2025deep}.} 
	\label{fig_indiv}
\end{figure*}

Based on these objective formulations, personalized exoskeleton assistance methods fall into two categories. Human-in-the-loop optimization (HILO) tunes user-specific assistance parameters using predefined measures, such as metabolic cost and preference feedback~\cite{zhang2017human,2018science,witte2020running,poggensee2021adaptation,lee2023user}. Data-fusion methods personalize assistance by aligning motion distributions across users or improving model generalization~\cite{liu2025weight,scherpereel2025deep,tan2026gaitdynamics}. Fig.~\ref{fig_indiv} schematically illustrates these two approaches, which are reviewed and compared in the following section.

\subsection{Human-in-the-loop Optimization}
Human-in-the-loop optimization (HILO) iteratively tunes control parameters through real-time human–exoskeleton interaction to meet predefined objectives \cite{liu2026pbbo}. The objective $\mathcal{J}$ may use physiological measures, including metabolic rate \cite{zhang2017human,2022TROHILO}, muscle activation \cite{musclereduce,2025knee}, and heart rate \cite{10102311}, or broader indicators such as user preference \cite{lee2023user}.
HILO requires sustained interaction, and repeated optimization cycles can cause fatigue, thereby limiting its use with patients. Moreover, tasks in complex environments yield high-dimensional assistance policies that are difficult to optimize directly when represented as neural networks. Exoskeleton HILO therefore faces two challenges:
\begin{itemize}[leftmargin=*]
    \item \textit{Challenge 3.1}: How can an efficient human-in-the-loop optimization algorithm reach a solution with limited human-robot interaction?
    \item \textit{Challenge 3.2}: How can a preference-optimization framework be developed for high-dimensional policy optimization on complex terrain?
\end{itemize}

\subsubsection{\textbf{Physiological-measure-based Optimization}}
Exoskeleton assistance aims to reduce metabolic cost and joint-injury risk~\cite{2018science, ingraham2022role}. Personalized optimization methods define the objective $\mathcal{J}$ using reductions in metabolic rate~\cite{2025hip,11606457}, muscle activation~\cite{2025knee,musclereduce}, or heart rate~\cite{10102311}; others combine multiple physiological measures into a composite objective~\cite{9992260,11075537}. During online assistance, physiological feedback guides the optimization of low-dimensional, task-specific parameters~\cite{zhang2017human}. Most studies tune torque-profile parameters, such as peak magnitude and timing, to individual gait characteristics and preferences~\cite{nuckols2021muscle}. Since prolonged human--robot interaction can cause fatigue, recent work has focused on identifying optimal parameters from fewer interaction trials.

Common algorithms are Bayesian optimization (BO)~\cite{2018science,2025hip,2025knee,arens2025preference} and covariance matrix adaptation evolution strategy (CMA-ES)~\cite{zhang2017human}. Both search a small parameter set $\omega$ online, with the state $\mathbf{x}$ given by assistance conditions such as walking speed and terrain slope. The parameters include torque-profile characteristics~\cite{2025hip,10546990,11112638} or impedance parameters~\cite{9815515} for personalized gait assistance. In~\cite{zhang2017human}, CMA-ES is used to minimize metabolic rate measured by respirometry. Gordon et al.~\cite{2022TROHILO} estimated metabolic rate online from a musculoskeletal model and applied BO, shortening the interaction and reducing fatigue. Fewer interaction trials can improve efficiency and address \textit{Challenge 3.1}. However, each evaluation still depends on sensing accuracy, and metabolic assessment requires a long collection window~\cite{2021preference,arens2025preference}, which lengthens experiments and increases uncertainty. Moreover, user experience is also influenced by fatigue, balance, and comfort, which cannot be fully captured by a single physiological measure~\cite{liu2026pbbo,lee2023user}. These limits have motivated preference-based personalization~\cite{slade2022personalizing,arens2025preference}, reviewed next.

\subsubsection{\textbf{Preference-based Optimization}}
User preference learning builds computational models to identify motion trajectories aligned with individual preferences and uses feedback to guide parameter optimization. Given two candidate curves $\sigma_1$ and $\sigma_2$, each generated from a parameter vector $\omega$, the user experiences both and provides comfort-based feedback $p$; $\sigma_1\succ\sigma_2$ indicates a preference for $\sigma_1$. These comparisons form a dataset $\mathcal{D}_{pre}$ of samples $(\sigma_1,\sigma_2,p)$, which guides optimization of personalized parameters $\omega$. Most methods follow two stages: learning an implicit objective $\mathcal{J}$ from preference feedback and then optimizing $\omega$ using that objective.

\textbullet \text{ } \textit{Implicit Objective Learning:} Assistance quality is difficult to express as an explicit mathematical objective. User feedback can instead be used to learn an implicit objective $\mathcal{J}$ that assigns a ranking score or value distribution to the current state $\mathbf{x}$. Existing methods fall into two groups: data-driven approaches \cite{slade2022personalizing,lee2023user} and statistical methods \cite{tucker2020preference,2021preference,arens2025preference}.
Data-driven methods train an evaluation network on an offline preference dataset to assess assistance quality. In \cite{slade2022personalizing}, a classifier was trained to estimate the probability that one control-parameter set is superior to another, enabling multiple configurations to be ranked. Using the Bradley–Terry model, Lee et al. \cite{lee2023user} trained RankNet to distinguish users’ motion preferences. Such networks either compare parameter sets to select a preferred configuration or assign a value to a configuration. They require preference data from multiple users, resulting in substantial collection costs.
Statistical methods \cite{10611497,9341416,schafer2026user,arens2025preference,liu2026pbbo} typically use Gaussian models to update the mean and variance of an implicit objective from preference feedback. In \cite{tucker2020preference,arens2025preference,liu2026pbbo}, Gaussian-process (GP) preference models were developed and refined through a finite number of online human-exoskeleton interactions until a stopping criterion was met. These methods use real-time interaction data and therefore require no offline preference dataset \cite{schafer2026user}. 
Because collecting preference data is costly and readily fatigues the user \cite{tucker2020preference,11128809}, efficiently combining the learned implicit objective with the optimization algorithm is essential for improving optimization efficiency.

\textbullet \text{ } \textit{Parameter Optimization:} 
Building on learned preference models, personalized methods commonly use BO~\cite{arens2025preference,liu2026pbbo} or CMA-ES~\cite{lee2023user}, with modifications to improve sample efficiency (\textit{Challenge 3.1}). A preference model was trained on offline data to quantify motion preferences and combined with CMA-ES to optimize parameters~\cite{lee2023user}. A kinematics-informed scoring model was updated through online interaction to tune control parameters~\cite{slade2022personalizing}. GP optimization is improved by prior knowledge of just-noticeable differences between assistance settings~\cite{arens2025preference}. On this basis, an adaptive sampling distribution that progressively narrows the search space and further improves optimization efficiency was developed in \cite{liu2026pbbo}. This method requires no offline dataset and optimizes parameters with limited online interaction.
However, these methods remain restricted to low-dimensional parameter spaces and specific conditions, such as fixed walking speeds and terrain slopes. Extending preference-based optimization to high-dimensional policy networks under complex conditions remains challenging (\textit{Challenge 3.2}).

\textbullet \text{ } \textit{Complex Policy Optimization:} 
Few exoskeleton studies optimize high-dimensional policies using preference feedback (\textit{Challenge 3.2}). The primary obstacle is the cost of collecting pairwise comparisons in a tightly coupled human-robot system, where applying two assistance profiles $\sigma_1$ and $\sigma_2$ under the same state $\mathbf{x}$ is difficult. In simulation or robotic manipulation, by contrast, states can be prescribed or reset, enabling comparisons under matched conditions.
Similarly, preference-based methods for optimizing complex policies have been proposed for robotic manipulation and for settings with realistic simulation. These methods may inform future optimization of exoskeleton assistance policies. Existing approaches fall into two categories. One learns a reward function from preference data to guide policy optimization \cite{LiangA-RSS-26,pmlr-v306-taranovic26a}, and is suited to settings in which the simulator is realistic and the policy is readily trainable. The other centers on designing an optimization objective that updates the policy directly from preference data \cite{NEURIPS2025_347110fb,cai2026predictive}. Such methods can guide the design of optimization frameworks for complex exoskeleton policies. The cost of collecting preference data, and the suitability of these methods for tightly coupled human-robot systems, should still be considered.

\begin{table}
    \normalsize
    \caption{Summary of representative human-in-the-loop optimization works for exoskeleton individual adaptation.}
    \label{tab_hilo}
    \centering
    \renewcommand{\arraystretch}{1}
    \setlength{\tabcolsep}{5pt}
    \arrayrulecolor{black}
    \resizebox{8.8cm}{!}{%
        \begin{tabular}{@{}l|ccl@{}}
            \toprule[1pt]
            \rowcolor{TableBlue}
            \multicolumn{4}{c}{\textbf{Physiological-measure-based Optimization}} \\
            {Related Works} & {Physiological Signal} & {Opt. Method} & {Parameters (Dim.)}\\
            \midrule[0.3pt]

            \textcolor{magenta}{\mbox{Powell} et al. (2026)} \cite{11606457} &  Metabolic Rate & BO & Delay, Shape (\textcolor{magenta}{2})\\
            \rowcolor{TableStripe}            
            \textcolor{magenta}{Gordon et al. (2022)} \cite{2022TROHILO} & Metabolic Rate & BO & Hip Profile (\textcolor{magenta}{4})\\
            \textcolor{magenta}{Zhang et al. (2017)} \cite{zhang2017human} & Metabolic Rate & CMA-ES & Ankle Profile (\textcolor{magenta}{4})\\
            
            \rowcolor{TableStripe}             
            \textcolor{magenta}{Wang et al. (2025)} \cite{2025knee} & Muscle Activation & BO & Knee Profile (\textcolor{magenta}{4})\\  
            \textcolor{magenta}{Zhang et al. (2025)} \cite{11075537} & Combined Signal & Genetic & Ankle Profile (\textcolor{magenta}{2})\\

            \rowcolor{TableStripe}     
            \textcolor{magenta}{Shafer et al. (2023)} \cite{9815515} & Metabolic Rate & - & Impedance (\textcolor{magenta}{2})\\        
            \midrule[0.7pt]
            \rowcolor{TableBlue}
            
            \multicolumn{4}{c}{\textbf{Preference-based Optimization}} \\
            {Related Works} & {Implicit Objective} & {Opt. Method} & Parameters (Dim.)\\
            \midrule[0.3pt]

            \textcolor{magenta}{Slade et al. (2022)} \cite{slade2022personalizing} & Data-Driven & CMA-ES &  Ankle Profile (\textcolor{magenta}{4})\\
            \rowcolor{TableStripe} 
            \textcolor{magenta}{Lee et al. (2023)} \cite{lee2023user} & Data-Driven & CMA-ES & Ankle Profile (\textcolor{magenta}{4})\\

            \textcolor{magenta}{Liu et al. (2026b)} \cite{liu2026pbbo} & Statistics-Based & BO &  Hip Profile (\textcolor{magenta}{6})\\
            \rowcolor{TableStripe}             
            \textcolor{magenta}{Arens et al. (2025)} \cite{arens2025preference} & Statistics-Based & BO &  Back Profile (\textcolor{magenta}{2})\\
        
            \textcolor{magenta}{Tucker et al. (2020)} \cite{tucker2020preference} & Statistics-Based & BO &   Hip Profile (\textcolor{magenta}{2})\\
            \bottomrule[1pt]
        \end{tabular}%
    }
\end{table}

\subsubsection{\textbf{Summary}}
Exoskeletons provide torque-based assistance, but optimal torque profiles vary across users even for the same task~\cite{slade2022personalizing}. When these profiles cannot be measured directly, human-in-the-loop optimization (HILO) can identify personalized assistance parameters~\cite{arens2025preference}. Existing methods rely on either physiological feedback, such as metabolic cost and muscle activation~\cite{zhang2017human,2022TROHILO,musclereduce,2025knee}, or user-preference feedback~\cite{lee2023user,arens2025preference,liu2026pbbo}. Preference feedback can capture individual motion characteristics and overall assistance quality more comprehensively. However, prolonged interaction may cause user fatigue~\cite{liu2026pbbo}, motivating more sample-efficient optimization with fewer trials (\textit{Challenge 3.1}).
Table~\ref{tab_hilo} compares representative methods in terms of their optimized parameters and optimization algorithms. Exoskeleton HILO remains largely restricted to small sets of key parameters~\cite{zhang2017human}, while optimization of high-dimensional network policies remains underexplored (\textit{Challenge 3.2}). Future research should develop efficient preference-based optimization methods for tightly coupled human-robot systems using preference data that can be collected easily and inexpensively.

\subsection{Cross-Subject Transfer}
Inter-subject differences in electromyographic activation, movement habits, and joint kinematics and dynamics create substantial shifts in multimodal data distributions \cite{scherpereel2025deep,bonyani2024user}. Models trained on one subject therefore generalize poorly to unseen subjects \cite{zhang2022gaussian,zhang2023ensemble}. An alternative approach exploits the structure of multimodal data to learn subject-invariant latent representations or align feature distributions across subjects. Existing studies follow two directions: learning shared representations for cross-subject knowledge transfer \cite{tu2026personalized,tan2026gaitdynamics}, and aligning source- and target-subject features for personalized prediction \cite{scherpereel2025deep,liu2025weight}. Together, they address the following challenge:
\begin{itemize}[leftmargin=*]
    \item \textit{Challenge 3.3}: How can the training objective be designed to improve generalization of personalized motion-intention prediction?
\end{itemize}

These methods pool data from multiple subjects and tasks to train a unified model $\mathcal{F}_\omega$ directly with the objective $\mathcal{J}$, eliminating separate task-adaptive models (Section \ref{sec_task}). In most studies, cross-subject motion-intention accuracy is reported through the accuracy of predicted torque or joint trajectories, or of gait recognition. Accurate recognition does not imply more effective assistance, and the relationship between the two warrants further investigation.

\subsubsection{\textbf{Multi-Subject Shared Representation Learning}} 
These methods depend on the quality, diversity, and scale of training data and typically follow two stages: learning shared representations and adapting to individuals. In the first stage, CNNs capture spatial correlations across channels and LSTMs model temporal dependencies \cite{lu2022continuous}, while TCN and TCN-BiLSTM architectures learn shared temporal patterns \cite{zhang2026generalized}. Preprocessing can further support personalization through muscle-synergy extraction using nonnegative matrix factorization (NMF) \cite{zhang2026generalized} or motion-pattern normalization \cite{fu2026effective}.

The second stage uses three strategies. First, high-capacity networks trained on large multi-subject datasets directly predict motion intentions for new users \cite{tan2026gaitdynamics,molinaro2024task,tu2026personalized}. Tan et al. \cite{tan2026gaitdynamics} generated gait motions and forces with a diffusion-based gait foundation model. Second, limited-data adaptation uses meta-learning across subjects \cite{cao2024fusion,zhang2026generalized}, with few-sample fine-tuning \cite{zhang2026generalized}, adversarial transfer on limited unlabeled target sEMG \cite{zhang2025pt}, or subject-specific adapters with memory replay to mitigate forgetting \cite{chen2024cross}. Third, physiology-guided methods \cite{tu2026personalized,fu2026effective} personalize EMG-based intention recognition using height, weight, and segment dimensions.
These three strategies have distinct limitations. The first relies on large, high-quality datasets; the second relies on loss-function design, which can make training unstable; and the third adapts poorly to complex, dynamic scenarios. The two stages are carried out either with a unified objective, often a regression loss such as mean squared error \cite{tan2026gaitdynamics,molinaro2024task}, or by fine-tuning the shared model with a subject-specific objective \cite{zhang2026generalized,chen2024cross}. Cross-subject generalization still depends on data richness and quality \cite{tan2026gaitdynamics}. Gait foundation models suggest further potential.

\subsubsection{\textbf{Cross-Subject Domain Adaptation}}
Domain adaptation learns task-discriminative features that generalize across domains from labeled data of source subjects or devices and limited labeled or unlabeled target data, reducing exoskeleton calibration costs \cite{liu2025weight,scherpereel2025deep,9446851}. For an unlabeled target, let the $m$-th labeled source domain be $\mathcal{D}_{sm} = \{(\mathbf{x}_{sm}^i, \mathbf{y}_{sm}^i)\}_{i=1}^{|\mathcal{D}_{sm}|}$, the $N$ source domains be $\mathcal{D}_{s}=\{\mathcal{D}_{sm}\}_{m=1}^N$, and the target be $\mathcal{D}_t = \{\mathbf{x}_t^i\}_{i=1}^{|\mathcal{D}_t|}$ \cite{gong2025multi,zhang2023ensemble}. Here $\mathbf{x}$ denotes human or exoskeleton sensor signals; $\mathbf{y}$ denotes motion mode, gait state, joint angle, or joint torque, with mapping $\mathcal{F}$ in Eq. \eqref{eq_3_1}. It is formulated as
\begin{equation}
  \min_{\omega} \text{ } \mathcal{J} = \sum_{m=1}^{N} \alpha_m \mathcal{L}_{\text{task}}^{m} + \beta_m \mathcal{L}_{\text{align}}^m + \gamma \mathcal{L}_{\text{cons}},
\end{equation}
where $\mathcal{L}_{\text{task}}^m$ is the classification cross-entropy or regression error on source $m$, $\mathcal{L}_{\text{align}}^m$ its discrepancy from the target, and $\mathcal{L}_{\text{cons}}$ an ideal-error constant that can be made small for a sufficiently expressive hypothesis space \cite{liu2025weight}. Coefficients $\alpha_m$ and $\beta_m$ weight source subjects, and source–target alignment improves target prediction.
Existing studies focus on two aspects: defining the source–target discrepancy $\mathcal{L}_{\text{align}}$ \cite{zhang2023ensemble,9446851,gong2025multi,liu2024robust} and assigning source-domain weights $\alpha$ and $\beta$ \cite{liu2025weight,hu2023swl,wen2024maximum} (\textit{Challenge 3.3}).

\textbullet \text{ } \textit{Discrepancy Determination:}
Domain adaptation minimizes the source–target discrepancy $\mathcal{L}_{\text{align}}$ to align features for target-domain prediction. Common measures include mean–variance matching \cite{zhang2022gaussian}, maximum mean discrepancy (MMD) \cite{9446851}, margin disparity discrepancy (MDD) \cite{liu2025weight}, and covariance alignment \cite{HAN2024109680}. Source and target means and variances are matched under a shared Gaussian \cite{zhang2022gaussian}. This matching is stable and non-adversarial, but misses higher-order shift. MMD combined with a regression contrastive loss was developed in \cite{9446851} for cross-subject sEMG wrist kinematics, so similar motions stay close and dissimilar ones separate; the contrastive term requires target labels. MDD measures how domain shift moves motion-intention classification margins \cite{liu2025weight}. This targets task-relevant shift, but applies to classification and needs auxiliary classifiers. An encoder then reduces these discrepancies for personalized prediction.

\textbullet \text{ } \textit{Weight Determination:}
Because source subjects differ in similarity to the target, weights $\alpha$ or $\beta$ scale their contributions by source–target discrepancy \cite{liu2025weight,gong2025multi,HAN2024109680} or by predictive reliability and information content \cite{xu2023dual,chakma2021activity}.
Discrepancy-based methods up-weight closer sources. Sources are weighted by Riemannian covariance distance \cite{HAN2024109680}, which is suited to EEG and multichannel sEMG. Similarity-weighted EEG classification was developed in \cite{gong2025multi}, giving each source a contribution according to its similarity to the target. Weights set from MDD also bound the target error \cite{liu2025weight}. These weights limit negative transfer, but the estimate can miss complex shifts, and min-max training may be unstable.
Reliability-based methods trust predictive evidence. Per-sample perplexity fusion of source classifiers \cite{chakma2021activity} and normalized mutual information of high-entropy samples combined with single-source validation accuracy \cite{xu2023dual} are used to determine the weights. Prediction replaces distribution distance, but confidence need not be correct without labels \cite{hu2023swl}, and validation accuracy needs labeled target data.
Combining discrepancy, uncertainty, and source error remains promising: discrepancy selects features to align, and weighting selects subjects to trust.

\begin{table}
    \normalsize
    \caption{Summary of representative cross-subject transfer works for exoskeleton individual adaptation.}
    \label{tab_cross}
    \centering
    \renewcommand{\arraystretch}{1}
    \setlength{\tabcolsep}{5pt}
    \arrayrulecolor{black}
    \resizebox{8.8cm}{!}{%
        \begin{tabular}{@{}l|ccc@{}}
            \toprule[1pt]
            \rowcolor{TableBlue}
            \multicolumn{4}{c}{\textbf{Multi-Subject Shared Representation}} \\
            {Related Works} & {Feature Encoder} & {Individual Adapter} & Task Type\\
            \midrule[0.3pt]
            \textcolor{magenta}{Molinaro et al. (2024a)} \cite{molinaro2024estimating} & TCN & -- &Torque\\
            \rowcolor{TableStripe}
            \textcolor{magenta}{Molinaro et al. (2024b)} \cite{molinaro2024task} & TCN & -- & Torque\\
            \textcolor{magenta}{Cao et al. (2024)} \cite{cao2024fusion} & MLP & Meta-Learning & Trajectory\\
            \rowcolor{TableStripe}
            \textcolor{magenta}{X. Chen et al. (2024)} \cite{chen2024cross} & TCN & Lifelong Learning & Angle\\
            \textcolor{magenta}{Tan et al. (2026)} \cite{tan2026gaitdynamics} & Transformer & -- & Gait Dynamics\\
                        
            \midrule[0.7pt]

            \rowcolor{TableBlue}
            \multicolumn{4}{c}{\textbf{Cross-Subject Domain Adaptation}} \\
            {Related Works} & {Discrepancy} & {Weight Determination} & Task Type\\
            \midrule[0.3pt]
            \textcolor{magenta}{Scherpereel et al. (2025)} \cite{scherpereel2025deep} & LS‑GAN & -- & Torque\\
            \rowcolor{TableStripe}
            \textcolor{magenta}{Liu et al. (2025)} \cite{liu2025weight} & MDD & Discrepancy-Based & Pattern Classif.\\
            \textcolor{magenta}{Gong et al. (2025)} \cite{gong2025multi} & MMD & Discrepancy-Based & EEG Classif.\\
            \rowcolor{TableStripe}
            \textcolor{magenta}{Xu et al. (2023)} \cite{xu2023dual} & MMD & Information-Based & EEG Classif.\\
            \textcolor{magenta}{Zhang et al. (2022)} \cite{zhang2022gaussian} & MVM & -- & Pattern Classif. \\
            \bottomrule[1pt]
        \end{tabular}%
    }
\end{table}

\subsubsection{\textbf{Summary}}
Cross-subject transfer for personalized exoskeleton intention prediction follows two directions: shared representation learning~\cite{molinaro2024task,tu2026personalized,tan2026gaitdynamics} and domain adaptation~\cite{scherpereel2025deep,liu2025weight} (Table~\ref{tab_cross}).
Shared-representation methods first train an encoder to extract common features from multi-subject data, and then train an adapter or decoder for new subjects. Performance depends on data quality and scale and is further supported by foundation models~\cite{tan2026gaitdynamics}. Domain-adaptation methods instead align source and target distributions by reducing discrepancies~\cite{zhang2022gaussian,bonyani2024user,liu2024robust} or weighting source domains~\cite{liu2025weight,gong2025multi} (\textit{Challenge 3.3}). They use smaller networks and yield more accurate personalized predictions than shared-representation methods, but require per-subject adaptation, which complicates deployment~\cite{liu2025weight,scherpereel2025deep}. Both directions learn transferable features from tailored objectives. High cross-subject prediction accuracy, however, does not imply effective assistance, which also depends on exoskeleton structure, control strategy, and other factors.

\subsection{Summary}
Differences in users’ motor patterns and preferences require exoskeletons to provide individualized assistance. Research follows two main approaches: human-in-the-loop optimization (HILO) of personalized parameters \cite{slade2022personalizing,lee2023user,arens2025preference} and cross-subject transfer for generalization to new users \cite{molinaro2024task,tan2026gaitdynamics,tu2026personalized}.
HILO methods use physiological measures, such as metabolic cost \cite{11606457} and muscle activation \cite{musclereduce}, or user-preference feedback \cite{liu2026pbbo,lee2023user} to guide optimization. They perform well in low-dimensional parameter spaces, but reducing online interaction and optimizing high-dimensional policies remain challenging \cite{liu2026pbbo}. Cross-subject transfer instead uses tailored objectives and network architectures to learn shared features \cite{tan2026gaitdynamics} and align data across users \cite{scherpereel2025deep}. Its success depends on sufficient high-quality data. Although both approaches have advanced personalized assistance-curve optimization and cross-subject prediction, optimizing complex policies for exoskeleton tasks remains an open problem.
    \section{Low-level Controller: Control Regulation} \label{Control}
Lower-limb exoskeletons typically employ a hierarchical control architecture, in which a high-level controller generates personalized, task-dependent references $\mathbf{y}$, while a low-level controller computes motor torques $\mathbf{u}$ to track the desired motion or torque profiles \cite{liu2026exotraj}. The low-level controller can be formulated as a reference-to-torque mapping:
\begin{equation}
    \mathbf{u}=\mathcal{G}(\mathbf{y}),
\end{equation}
where $\mathcal{G}(\cdot)$ is defined either analytically or implicitly through optimization. However, voluntary human motion, transmission compliance, and foot–ground contact transitions introduce substantial uncertainty and variability into the coupled human–exoskeleton dynamics \cite{li2022impedance}, making it difficult to model the system accurately and to guarantee tracking performance and interaction safety. Therefore, low-level control faces two key challenges:
\begin{itemize}[leftmargin=*]
\item \textit{Challenge 4.1}: How can accurate command tracking be achieved under uncertain dynamics, including actuation nonlinearities and contact transitions?
\item \textit{Challenge 4.2}: How can the assistance objectives be formulated to balance task performance, interaction safety, and the user's effort and voluntary motion?
\end{itemize}

\begin{figure}
	\centering
	\includegraphics[width=0.5\textwidth]{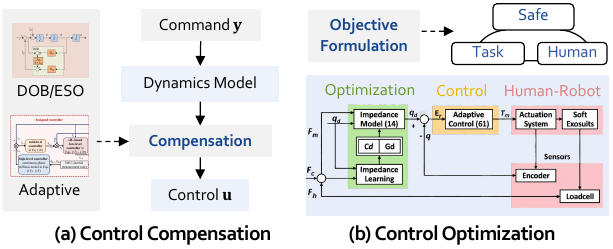}
	\caption{Schematic of two approaches to self-regulation control. (a) Control compensation: Dynamics-based feedforward compensation is combined with either disturbance-observer (DOB)-based compensation \cite{mohammadi2017nonlinear} or adaptive iterative compensation \cite{xing2024spatial} to achieve accurate tracking. (b) Control optimization: Key control parameters or torques are optimized based on safety, comfort, and human-movement metrics to enhance effectiveness \cite{li2022impedance}.} 
	\label{fig_con}
\end{figure}

Fig. \ref{fig_con} illustrates two primary approaches to self-regulation control. The first, control compensation, uses dynamic models to compensate for uncertainties and achieve accurate command tracking (\textit{Challenge 4.1}). The second, control optimization, formulates suitable objectives to ensure task completion while preserving assistance safety and comfort (\textit{Challenge 4.2}). Existing studies are reviewed below from these two perspectives.

\subsection{Control Compensation}

Low-level control of powered lower-limb exoskeletons regulates actuator outputs $\mathbf{u}$ to track prescribed joint trajectories or torque profiles $\mathbf{y}$~\cite{sharifi2022trajectory}. Under ideal torque actuation, the system dynamics can be described as~\cite{sharifi2022trajectory}:
\begin{equation}
    \mathrm{M}(\mathrm{q})\ddot{\mathrm{q}}
    + \mathrm{C}(\mathrm{q},\dot{\mathrm{q}})\dot{\mathrm{q}}
    + \mathrm{G}(\mathrm{q})
    + \mathrm{T}_{\mathrm{fric}}
    = \mathrm{T} + \mathrm{T}_{\mathrm{int}},
\end{equation}
where $\mathrm{q}$ and $\mathrm{T}$ denote the joint-angle and motor-torque vectors, respectively ($\mathrm{T}$ is $\mathbf{u}$). $\mathrm{M}$, $\mathrm{C}$, and $\mathrm{G}$ represent the inertia matrix, Coriolis matrix, and gravity vector, while $\mathrm{T}_{\mathrm{fric}}$ and $\mathrm{T}_{\mathrm{int}}$ denote the friction and wearer–exoskeleton interaction torques, respectively~\cite{xing2024spatial}. However, friction, structural compliance, and strong human–robot coupling introduce significant dynamic uncertainties~\cite{li2022impedance} (\textit{Challenge 4.1}). The low-level control law (mapping function $\mathcal{G}$) can be functionally decomposed as follows:
\begin{equation}
    \mathrm{T}
    = \mathrm{T}_{\mathrm{nom}}
    + \mathrm{T}_{\mathrm{fd}}
    + \mathrm{T}_{\mathrm{learn}}
    - \widehat{\mathrm{T}}_{\mathrm{dist}},
    \label{eq:torque_decomposition}
\end{equation}
where $\mathrm{T}_{\mathrm{nom}}$ provides baseline control based on the nominal model; $\mathrm{T}_{\mathrm{fd}}$ corrects instantaneous tracking errors; $\mathrm{T}_{\mathrm{learn}}$ leverages historical data to compensate for repeatable or predictable residual errors; and $\widehat{\mathrm{T}}_{\mathrm{dist}}$ estimates the equivalent disturbance torque for rejection. Accordingly, compensation strategies are classified into two broad categories: i) model-based feedforward and observer-based compensation~\cite{kim2025improved,sharifi2022trajectory}, and ii) adaptive and learning-based compensation~\cite{li2022impedance,xing2024spatial}.

\subsubsection{\textbf{Feedforward and Observer-Based Compensation}}

Model-based feedforward generates $\mathrm{T}_{\mathrm{nom}}$ from identifiable dynamics~\cite{kim2017modelinverse,chen2018pediatriclookup,kim2025improved,orekhov2020regression}, and observers provide $-\widehat{\mathrm{T}}_{\mathrm{dist}}$ to reject residual input-equivalent disturbances left by the nominal model~\cite{chen2023fteso,sharifi2022trajectory}. Together with feedback $\mathrm{T}_{\mathrm{fd}}$, they compensate for transmission compliance, wearable-interface deformation, and uncertain human–exoskeleton coupling during prescribed trajectory or torque tracking.

\textbullet\enspace
\textit{Model-Based Feedforward Compensation:}
Feedforward control computes $\mathrm{T}_{\mathrm{nom}}$ from desired outputs using input-output models to compensate for predictable dynamics, whereas stabilizing feedback $\mathrm{T}_{\mathrm{fd}}$ corrects instantaneous tracking errors~\cite{lee2017assistive}.
Physics-based feedforward derives $\mathrm{T}_{\mathrm{nom}}$ from system dynamics~\cite{lee2017assistive,kim2017modelinverse}. Compliance and transmission models~\cite{lee2017assistive}, as well as inverse-reference correction~\cite{kim2017modelinverse}, have been used to determine $\mathrm{T}_{\mathrm{nom}}$. Reliable transmission identification enables physically interpretable compensation; however, modeling errors and variable human loading can degrade accuracy.
Empirical feedforward determines $\mathrm{T}_{\mathrm{nom}}$ from offline-calibrated mappings between desired outputs and control inputs~\cite{orekhov2020regression,chen2018pediatriclookup,kim2025improved}. Representative approaches include motor-current regression~\cite{orekhov2020regression} and phase-specific stiffness maps~\cite{kim2025improved}. These methods are well suited to complex transmissions when calibration data are available, but accuracy depends on data coverage, so changes in operating conditions may require recalibration.

\textbullet\enspace
\textit{Disturbance Observer-Based Compensation:}
Disturbance compensation \(\widehat{\mathrm{T}}_{\mathrm{dist}}\) captures residual friction, dynamic-model mismatch, and task-specific interactions~\cite{zhang2023leso}. Disturbance observers (DOBs)~\cite{chen2024learning,sharifi2022trajectory,park2023hybriddob} and extended state observers (ESOs)~\cite{9737684,zhang2023leso} reconstruct these disturbances online.
DOBs estimate the equivalent disturbance from system dynamics using nonlinear estimation and adaptive gains~\cite{park2023hybriddob}. Various designs~\cite{durandau2022neuromechanical,huo2020force,sharifi2022trajectory,park2023hybriddob} improve estimation speed and accuracy. However, their reliance on accurate models makes them susceptible to modeling errors, potentially degrading stability and increasing noise sensitivity.
ESOs augment the system state with lumped disturbances, jointly estimating \(\widehat{\mathrm{T}}_{\mathrm{dist}}\), unmeasured motion states, and system uncertainties~\cite{zhang2023leso}. Finite-time ESOs~\cite{chen2023fteso,wang2022observer} can reduce sensing requirements without explicit disturbance models. However, their performance depends on system observability, while high observer gains may amplify measurement noise.

\subsubsection{\textbf{Adaptive and Learning-Based Compensation}}

Adaptive nonlinear approximation~\cite{li2022impedance,sun2021fuzzy,li2025fuzzy} and cross-stride iterative learning~\cite{yang2022spatialpath,xia2023spatialguidance,xing2024spatial} construct $\mathrm{T}_{\mathrm{learn}}$ using tracking errors and data from previous executions~\cite{xing2024spatial}. Unlike the fixed terms $\mathrm{T}_{\mathrm{nom}}$ and $\mathrm{T}_{\mathrm{fd}}$, $\mathrm{T}_{\mathrm{learn}}$ encompasses updates to model parameters, controller gains, and nonlinear approximators, thereby adapting to variations in human capability and actuator dynamics as well as systematic recurring errors~\cite{li2022impedance}.

\textbullet\enspace
\textit{Adaptive Updates and Fuzzy Approximation:}
Robust parameter adaptation and fuzzy approximation update \(\mathrm{T}_{\mathrm{learn}}\) online based on the instantaneous tracking error, while the robust feedback term \(\mathrm{T}_{\mathrm{fd}}\) suppresses residual uncertainties~\cite{li2025fuzzy}.
Robust parameter adaptation exploits known dynamic properties or disturbance bounds~\cite{chen2017hydraulic,li2023adaptation,li2022impedance} to modify the nominal and feedback control laws. These modifications are incorporated into \(\mathrm{T}_{\mathrm{learn}}\) as incremental corrections to the fixed terms \(\mathrm{T}_{\mathrm{nom}}\) and \(\mathrm{T}_{\mathrm{fd}}\). Such methods are well suited to structured or bounded uncertainties; however, excessively large gains may induce chattering.
Fuzzy approximation methods, such as reduced-order fuzzy compensators~\cite{sun2021fuzzy} and saturation-aware compensators~\cite{li2025fuzzy}, characterize unknown nonlinearities through state-dependent rules and basis functions with online parameter adaptation. The resulting approximation constitutes \(\mathrm{T}_{\mathrm{learn}}\) and reduces the reliance on explicit nonlinear models. Nevertheless, their performance depends strongly on the design of the rule base and on the adaptation laws.

\textbullet\enspace
\textit{Cross-Stride Iterative Learning Compensation:}
Iterative learning updates subsequent actuator commands $\mathbf{u}$ using errors from preceding gait cycles, converting repeatable deviations into feedforward compensation $\mathrm{T}_{\mathrm{learn}}$ indexed in the time/gait-phase domain~\cite{chen2024ankle} or the spatial domain~\cite{xing2024spatial}.
Time/gait-phase learning aligns errors across repetitive walking cycles according to comparable gait phases~\cite{bryan2021hip,tan2022iterative,chen2024ankle}. Feedback-assisted iterative updates progressively refine $\mathrm{T}_{\mathrm{learn}}$. However, this approach requires inter-stride repeatability and accurate phase alignment~\cite{livolsi2022wavelet}; abrupt changes in walking speed or assistance timing may invalidate the learned compensation.
Spatial learning indexes both tracking errors and $\mathrm{T}_{\mathrm{learn}}$ using spatial variables, thereby reducing temporal misalignment caused by speed variations~\cite{yang2022spatialpath,xia2023spatialguidance,xing2024spatial}. Although spatial learning relaxes the requirement for a fixed gait period, it assumes that tracking errors remain approximately repeatable at corresponding spatial states across successive cycles.

\subsubsection{\textbf{Summary}}
Addressing time-varying human-exoskeleton coupling, actuator-transmission nonlinearities, and contact-state transitions requires accurate real-time tracking (\textit{Challenge 4.1}) through two complementary families of compensation strategies. The first, model-based feedforward and observer-based compensation, combines physics-based or empirically identified nominal feedforward torque $\mathrm{T}_{\mathrm{nom}}$ with DOB/ESO-based compensation $-\widehat{\mathrm{T}}_{\mathrm{dist}}$ to reject residual uncertainties~\cite{orekhov2020regression}. The second, adaptive and learning-based compensation, constructs $\mathrm{T}_{\mathrm{learn}}$ through online parameter adaptation, nonlinear function approximation, and cross-stride iterative learning in the time/gait-phase or spatial domain~\cite{li2022impedance,sun2021fuzzy,xing2024spatial}. In both families, a stabilizing feedback term $\mathrm{T}_{\mathrm{fd}}$ corrects residual tracking errors. Overall performance depends on model fidelity, observer bandwidth, adaptation transients, and error repeatability.

\subsection{Control Optimization}
Addressing \textit{Challenge 4.2} requires preserving voluntary motion permitted by the task while providing sufficient assistance within safety and physical constraints~\cite{huo2022impedance}. Accurate tracking alone may suppress voluntary adjustments, whereas low interaction forces may indicate insufficient support~\cite{li2017multi,kuccuktabak2024haptic}. Because the required level of assistance is task-dependent, it should be defined through explicit performance criteria. Given high-level trajectory or torque references $\mathbf{y}$, the following formulation captures complementary objectives:
\begin{equation}
    \min_{\mathrm{z}\in\mathcal{Z}}
    C_t\mathcal{J}_{\mathrm{task}}(\mathrm{z})
    +C_s\mathcal{J}_{\mathrm{safe}}(\mathrm{z})
    +C_h\mathcal{J}_{\mathrm{human}}(\mathrm{z}),
\end{equation}
where $\mathrm{z}$ denotes the interaction parameters or control decisions $\mathbf{u}$, and $\mathcal{Z}$ encodes the system dynamics and explicit constraints on actuator outputs and stability. The objective $\mathcal{J}_{\mathrm{task}}$ quantifies task deviation or interaction mismatch, $\mathcal{J}_{\mathrm{safe}}$ penalizes constraint relaxation or safety-related risks, and $\mathcal{J}_{\mathrm{human}}$ evaluates human effort or behavioral disturbance. The nonnegative weights $C_t$, $C_s$, and $C_h$ determine the relative importance of these objectives. This formulation serves as a review-level abstraction rather than a universal objective shared by all studies.
Two complementary branches are considered: impedance interaction optimization~\cite{li2017multi,li2022impedance}, which specifies desired torque-motion relationships, and constrained assistance optimization~\cite{walters2026optimal,xu2025robust}, which determines feasible control outputs subject to the constraint space $\mathcal{Z}$.

\subsubsection{\textbf{Impedance Interaction Optimization}}
Impedance optimization adjusts stiffness, damping, or their parameterizations to shape the torque-motion relationship, correcting reference deviations while remaining compliant during interaction~\cite{zhong2025hierarchical}. The objectives are interaction matching, task-error minimization, and effort reduction.
Weighted stiffness and damping errors are minimized under the impedance residual and bounds \cite{li2022impedance}, so assistance adapts across terrains, subject to sensing accuracy, the reference, and the weights. Model-free path-integral policy improvement was developed in \cite{wang2024rm} to reduce trajectory error in walking, squatting, and climbing, though it remains limited by the reference and sampled rollouts. Impedance is initialized and then adapted using the estimated knee torque \cite{meng2024optimized} when that estimate is accurate. Extremum seeking on three-stride EMG tunes quasi-passive ankle stiffness \cite{kumar2020extremum}, with the search limited by noise and aggregation in slow walking.
This suits interaction that varies with terrain, task, or effort: force matching for changing contact and effort tuning for variable stiffness. Performance depends on the reference, when one is used, and on force, torque, or EMG accuracy.

\subsubsection{\textbf{Constrained Assistance Optimization}}
This branch optimizes assistive torque or controller parameters~\cite{yu2024safety}, with objectives that distinguish between preserving voluntary behavior and delivering the requested assistance~\cite{kuccuktabak2024haptic}. Hard, non-negotiable limits are encoded in the feasible set $\mathcal{Z}$, whereas $\mathcal{J}_{\mathrm{safe}}$ penalizes allowable constraint relaxation or protection-related objectives without replacing these hard constraints.

\begin{table}[!t]
    \normalsize
    \caption{Representative low-level control methods for lower-limb exoskeletons.}
    \label{tab_low_level_control}
    \centering
    \renewcommand{\arraystretch}{1}
    \setlength{\tabcolsep}{5pt}
    \arrayrulecolor{black}
    \resizebox{8.8cm}{!}{%
        \begin{tabular}{@{}l|ccc@{}}
            \toprule[1pt]
            \rowcolor{TableBlue}
            \multicolumn{4}{c}{\textbf{Control Compensation}} \\
            Related Works & Compensation & Issue & Target \\
            \midrule[0.3pt]

            \textcolor{magenta}{Sharifi (2022)}
            ~\cite{sharifi2022trajectory} &
            DOB ($\widehat{\mathrm{T}}_{\mathrm{dist}}$) &
            Interact. Torque & Trajectory \\

            \rowcolor{TableStripe}
            \textcolor{magenta}{Chen (2023)}
            ~\cite{chen2023fteso} &
            ESO ($\widehat{\mathrm{T}}_{\mathrm{dist}}$) &
            Lumped Uncert. & Trajectory \\

            \textcolor{magenta}{Li (2023)}
            ~\cite{li2023adaptation} &
            Adaptation ($\mathrm{T}_{\mathrm{learn}}$) &
            Dynamics & Torque \\

            \rowcolor{TableStripe}
            \textcolor{magenta}{Li (2025)}
            ~\cite{li2025fuzzy} &
            Fuzzy ($\mathrm{T}_{\mathrm{learn}}$) &
            Saturation & Trajectory \\

            \textcolor{magenta}{Xing (2024)}
            ~\cite{xing2024spatial} &
            Iterative Learn. ($\mathrm{T}_{\mathrm{learn}}$) &
            Spatial Uncert. & Torque \\

            \midrule[0.7pt]
            \rowcolor{TableBlue}
            \multicolumn{4}{c}{\textbf{Control Optimization}} \\
            Related Works & Objective & Variables & Stability/Constraints \\
            \midrule[0.3pt]

            \textcolor{magenta}{Huo (2022)}
            ~\cite{huo2022impedance} &
            Effort Reduction & Impedance & Lyapunov Analysis \\

            \rowcolor{TableStripe}
            \textcolor{magenta}{Li (2022)}
            ~\cite{li2022impedance} &
            Interaction Match. & Impedance & Parameter Bounds \\

            \textcolor{magenta}{Yang (2025)}
            ~\cite{yang2025multicontact} &
            Safe Adaptation & Impedance & CBF \\

            \rowcolor{TableStripe}
            \textcolor{magenta}{K{\"u}{\c{c}}{\"u}ktabak (2024)}
            ~\cite{kuccuktabak2024haptic} &
            Admittance Track. & Torques & Actuator Limits \\

            \textcolor{magenta}{Walters (2026)}
            ~\cite{walters2026optimal} &
            Biomimetic Assist. & Shape Coeffs. & Cond. Stability \\

            \bottomrule[1pt]
        \end{tabular}%
    }
\end{table}

\textbullet\enspace \textit{Safety Intervention that Preserves Voluntary Motion:}
Subject to the selected safety constraints, this objective minimizes unnecessary intervention: $\mathcal{J}_{\mathrm{human}}$ quantifies disturbance to the user's voluntary motion, whereas $\mathcal{J}_{\mathrm{task}}$ measures deviation from the nominal task command \cite{yu2024safety}. Complementary boundary-enforcement mechanisms have also been developed. Voluntary-motion regions are combined with corrective guidance \cite{li2017multi}. Knee-motion boundaries are constructed from hip-dependent probabilistic predictions \cite{chen2023fteso}. Modulation of assistance near trajectory-tube boundaries by control barrier functions (CBFs) was developed in \cite{gurriet2019towards}. CBF constraints are incorporated into impedance-based reference adaptation across terrains \cite{yang2025multicontact}. Such optimization problems rely on accurate dynamic models and a principled, objective formulation of the human cost function \(\mathcal{J}_{\mathrm{human}}\).

\textbullet\enspace \textit{Task-oriented Assistance under Constraints:}
These approaches optimize $\mathcal{J}_{\mathrm{task}}$ over the feasible set $\mathcal{Z}$, seeking sufficient assistance while respecting safety and actuation limits~\cite{liu2026exotraj}. Motor torques are optimized to realize admittance-derived acceleration commands subject to system dynamics \cite{kuccuktabak2024haptic}. Tracking and torque costs are combined under motion and input constraints \cite{caulcrick2021model}, and the tracking weights are reduced in response to human resistance. Integration of robust MPC with a DOB was developed in \cite{xu2025robust} to improve reference execution under uncertainty. Energy-shaping coefficients are optimized to provide biomimetic assistance across tasks \cite{lin2024modular,walters2026optimal}. By reconciling requested support with device capabilities, these approaches improve the realizability of assistance. However, inaccuracies in dynamic models can degrade optimization performance.

\begin{figure}
	\centering
	\includegraphics[width=0.5\textwidth]{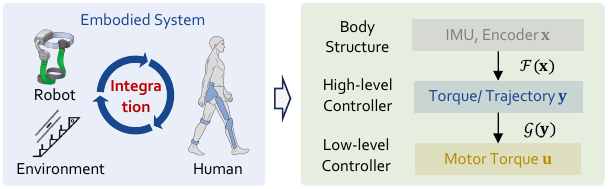}
	\caption{Schematic of exoskeleton assistance provided by the coupled human–robot–environment system. The high-level controller learns the mapping $\mathcal{F}$ from sensory information $\mathbf{x}$ to the desired trajectory or torque command \(\mathbf{y}\), while the low-level controller learns the mapping \(\mathcal{G}\) from the high-level command \(\mathbf{y}\) to the motor torque \(\mathbf{u}\).} 
	\label{fig_impl}
\end{figure}

\subsubsection{\textbf{Summary}}
Addressing \textit{Challenge 4.2} requires preserving voluntary motion and ensuring safety while delivering sufficient assistance within physical limits. Impedance optimization combines interaction-matching and task-error objectives for corrective guidance with effort-assistance matching to reduce muscular demand~\cite{li2022impedance,dos2019optimal,kumar2020extremum}. Constrained assistance optimization distinguishes autonomy-preserving intervention, which minimizes unnecessary behavioral or command changes, from task-oriented assistance, which realizes the requested support within the device's physical capabilities~\cite{yu2024safety,kuccuktabak2024haptic,walters2026optimal}. Together, these branches integrate interaction shaping with feasible assistance delivery, although their performance depends on dynamics model accuracy and constraint feasibility.

\subsection{Summary}
Given a high-level command \(\mathbf{y}\), the low-level controller generates a control input \(\mathbf{u}\) by either leveraging prior physical knowledge or optimizing \(\mathbf{u}\) using the prescribed objective. Control compensation addresses \textit{Challenge 4.1} by combining model-based feedforward, disturbance observation, online adaptation, and iterative learning with stabilizing feedback, thereby improving command tracking for uncertain coupled human–exoskeleton systems~\cite{li2022impedance,xing2024spatial}. For \textit{Challenge 4.2}, impedance optimization tunes interaction parameters to balance corrective guidance with human effort, whereas constrained optimization selects feasible outputs to preserve voluntary behavior and ensure safety while delivering the required assistance~\cite{dos2019optimal,yu2024safety,kuccuktabak2024haptic}.
Table~\ref{tab_low_level_control} summarizes representative studies. For control compensation, it compares compensation methods, key challenges, and control targets; for control optimization, it compares objectives, tuning variables, and stability conditions or constraints. Together, these approaches coordinate tracking accuracy, compliant interaction, and assistance sufficiency.
Nevertheless, these methods remain sensitive to model accuracy. Future research could explore data-driven approaches to learn high-fidelity human–robot dynamics models in complex scenarios.

    \section{Actual Implementation}\label{sec_imple}

\begin{table*}[t]
    \normalsize
    \caption{Comparative analysis of hardware systems, high-/low-level control, experimental scenarios, and performance across state-of-the-art lower-limb exoskeleton studies.}
    \label{tab:multitask_exoskeletons}
    \centering
    \renewcommand{\arraystretch}{1.12}
    \setlength{\tabcolsep}{3pt}
    \arrayrulecolor{black}
    \resizebox{18.0cm}{!}{%
        \begin{tabular}{@{}l|ccc|cc|cc|cc@{}}
            \toprule[1pt]
            \rowcolor{TableSubHeader}
            \cellcolor{TableStubHeader}
            & \multicolumn{3}{c!{\vrule width0.5pt}}{\cellcolor{TableHardwareHeader}\textbf{Hardware}}
            & \multicolumn{2}{c!{\vrule width0.5pt}}{\cellcolor{TableHighHeader}\textbf{High-Level Controller}}
            & \multicolumn{2}{c!{\vrule width0.5pt}}{\cellcolor{TableLowHeader}\textbf{Low-Level Controller}}
            & \multicolumn{2}{c}{\cellcolor{TableResultsHeader}\textbf{Experimental Results}} \\
            \rowcolor{TableSubHeader}
            \cellcolor{TableStubHeader}\multirow{-2}{*}{\textbf{Related Works}} &
            \textbf{Joint} & \textbf{Mass (kg)}\textsuperscript{c} & \textbf{Sensors} &
            \textbf{Algorithm} & \textbf{Output} &
            \textbf{Method} & \textbf{Cmd. (Hz)}\textsuperscript{a} &
            \textbf{Tasks/Terrains} & \textbf{Reduction (\%)}\textsuperscript{b} \\
            \midrule[0.4pt]

            \textcolor{magenta}{Kim (2019)} \cite{kim2019reducing} &
            Hip & 5.0 & IMU & CoM-Based Rules & Force &
            -- & -- & Level/Slope Walk, Run & 9.3 (n=9) \\

            \rowcolor{TableStripe}
            \textcolor{magenta}{Slade (2022)} \cite{slade2022personalizing} &
            Ankle & 2.7 & Enc., PI, SG & LR + CMA-ES & Curve Params. &
            Feedback + ILC & 200 & Level/Slope Walk & 23.0 (n=10) \\

            \textcolor{magenta}{Luo (2024)} \cite{luo2024experiment} &
            Hip & 3.2 & IMU, Enc. & MLP (RL) & Angle Ref. &
            PD + Current Ctrl. & 100 & Walk, Run, Stairs & 24.3 (n=8) \\

            \rowcolor{TableStripe}
            \textcolor{magenta}{Molinaro (2024b)} \cite{molinaro2024task} &
            Hip-Knee & 7.0 & IMU, Enc., PI & TCN & Torque &
            -- & 55 & Walk/Run, Lift, Lunge & -8.0 (n=9)\\

            \textcolor{magenta}{Molinaro (2024a)} \cite{molinaro2024estimating} &
            Hip & 4.8 & IMU, Enc. & TCN & Torque &
            Current Feedback & 200 & Level, Ramps, Stairs & 5.4 (n=10) \\

            \rowcolor{TableStripe}
            \textcolor{magenta}{Tricomi (2024)} \cite{tricomi2024soft} &
            Hip & 2.9 & IMU, Enc. & Phase Portrait & Gait Phase &
            Position PD & 100 & Level, Hills, Steps & 10.5 (Off, n=10, old) \\

            \textcolor{magenta}{Tricomi (2025)} \cite{tricomi2025leveraging} &
            Hip & 2.7 & IMU, Enc., RGB & GM + AO & Terrain/Phase &
            Position PID & 100 & Level, Stairs & 13.4 (Off, n=6) \\

            \rowcolor{TableStripe}
            \textcolor{magenta}{Durandau (2022)} \cite{durandau2022neuromechanical} &
            Ankle & 20 & sEMG, Enc. & EMG-MSK Model & Torque &
            PD + FF + DOB & 1000 & Var. Speed/Grade Walk & -- \\

            \textcolor{magenta}{Chen (2024)} \cite{chen2024learning} &
            Knee & -- & Enc., SG & MLP/VAE + DMP & Angle Traj. &
            DOB + VIC & -- & Walk, Stairs, Squat & -- \\

            \rowcolor{TableStripe}
            \textcolor{magenta}{Medrano (2023)} \cite{medrano2023realtime} &
            Ankle & -- & IMU & EKF & Gait Params. &
            Current PID & -- & Uneven/Sloped Walk & -- \\

            \textcolor{magenta}{Lee (2024)} \cite{lee2024universal} &
            Knee & 4.4 & IMU, Enc. & CNN & Gait Params. &
            Pseudo-Impedance & 100 & Level, Ramps, Stairs & 6.2 (Off, n=10) \\

            \rowcolor{TableStripe}
            \textcolor{magenta}{X. Zhang (2026)} \cite{zhang2026enhancing} &
            Hip & 2.9 & IMU, Enc. & RF + FSM & Task/Phase &
            P + LPF & -- & Level Walk, STS & 18.0 (n=7, old) \\

            \textcolor{magenta}{Y. Zhang (2026)} \cite{zhang2026data} &
            Hip & 4.0 & IMU, Enc. & CNN + AO & Mode/Phase &
            -- & 200 & Level, Uphill, Stair Ascent & 15.3 (n=5) \\

            \bottomrule[1pt]
        \end{tabular}%
    }
    \par\vspace{2pt}
    \begin{minipage}{18.0cm}
        \scriptsize
        \textsuperscript{a} Assistance-command update rate. Inner motor-current loops are not listed.
        \textsuperscript{b} Default: level-walking net metabolic reduction versus walking without the exoskeleton (No Exo). (Off): assistance off. The parenthetical \(n\) is the number of participants in that comparison. (old) denotes older adults; cohorts without this mark are young adults.
        \textsuperscript{c} Worn mass, including the battery and controller. An asterisk marks a reported exoskeleton mass that excludes those components.

        \textit{Abbreviations:}
        Enc., Encoder; PI, Pressure Insole; SG, Strain Gauge;
        CoM, Center of Mass;
        LR, Logistic Regression;
        MLP, Multilayer Perceptron;
        VAE, Variational Autoencoder; EKF, Extended Kalman Filter;
        RF, Random Forest; FSM, Finite-State Machine; AO, Adaptive Oscillator; GM, Geometric Model; FM, Flow Matching;
        MSK, Musculoskeletal;
        FF, Feedforward; VIC, Variable Impedance Control; ILC, Iterative Learning Control;
        LPF, Low-Pass Filter; STS, Sit-to-Stand; COT, Cost of Transport.
    \end{minipage}
\end{table*}

\begin{table}[!t]
    \normalsize
    \caption{Hardware models of sensing and control subsystems in representative lower-limb exoskeleton studies.}
    \label{tab:exoskeleton_hardware}
    \centering
    \renewcommand{\arraystretch}{1.12}
    \setlength{\tabcolsep}{5pt}
    \arrayrulecolor{black}
    \resizebox{8.8cm}{!}{%
        \begin{tabular}{@{}l|cccc@{}}
            \toprule[1pt]
            \rowcolor{TableBlue}
            \textbf{Related Works} &
            \textbf{IMU} &
            \textbf{Motor} &
            \textbf{High-Level} &
            \textbf{Low-Level} \\
            \midrule[0.4pt]

            \textcolor{magenta}{Kim (2019)} \cite{kim2019reducing} &
            MTi-3 AHRS & Maxon 305013 & ATSAME70N21 & Gold Twitter \\

            \rowcolor{TableStripe}
            \textcolor{magenta}{Slade (2022)} \cite{slade2022personalizing} &
            -- & AK80-9 & RPi 4B & RPi 4B \\

            \textcolor{magenta}{Luo (2024)} \cite{luo2024experiment} &
            LPMS-B2 & RMD-X8 & RPi 4 & Teensy 4.1 \\

            \rowcolor{TableStripe}
            \textcolor{magenta}{Molinaro (2024b)} \cite{molinaro2024task} &
            OpenIMU & AK80-9 & Jetson Nano & RPi 4B \\

            \textcolor{magenta}{Molinaro (2024a)} \cite{molinaro2024estimating} &
            MPU-9250 & AK80-9 & Jetson Nano & myRIO-1900 \\

            \textcolor{magenta}{Tricomi (2025)} \cite{tricomi2025leveraging} &
            BNO055 & AK60-6 & Jetson Nano & Arduino MKR \\

            \rowcolor{TableStripe}
            \textcolor{magenta}{Durandau (2022)} \cite{durandau2022neuromechanical} &
            Xsens Link & U8-10 Pro & Intel NUC & Intel NUC \\

            \textcolor{magenta}{Lee (2024)} \cite{lee2024universal} &
            MPU-9250 & AK80-9 & Jetson Nano & NI myRIO \\

            \rowcolor{TableStripe}
            \textcolor{magenta}{X. Zhang (2026)} \cite{zhang2026enhancing} &
            BNO055 & AK60-6 & Jetson Nano & Arduino MKR \\

            \textcolor{magenta}{Y. Zhang (2026)} \cite{zhang2026data} &
            JY901B & GIM8018-9 & Jetson Orin NX & STM32H723 \\

            \bottomrule[1pt]
        \end{tabular}%
    }
\end{table}

Lower-limb exoskeletons aim to reduce metabolic demand and joint loading across diverse tasks and terrains while accommodating coupled human-robot-environment dynamics~\cite{molinaro2024task,schmitz2022achilles}. Fig.~\ref{fig_impl} schematically illustrates assistance within this coupled system.
Effective assistance requires lightweight, biomechanically compatible hardware and hierarchical control tailored to individual users and task demands~\cite{luo2023robust,slade2022personalizing}. High-level controllers infer movement intent and generate personalized assistance references, whereas low-level controllers regulate actuator outputs to deliver robust real-time assistance~\cite{liu2026exotraj}. This section reviews hardware design, adaptive control at both levels, and experimental performance.

\subsection{System Configurations}
Exoskeleton electronics integrate sensing, onboard computation and control, power management, and communication~\cite{williams2025openexo}. Table~\ref{tab:exoskeleton_hardware} summarizes representative sensing, actuation, and computing hardware used in lower-limb exoskeletons.
Inertial measurement units (IMUs) and motor-integrated encoders commonly provide kinematic feedback. Frequently used IMUs include the MPU-9250~\cite{liu2026exotraj}, BNO055~\cite{zhang2026enhancing}, and OpenIMU series~\cite{molinaro2024task}, while AK-series actuators are widely adopted in the reviewed systems~\cite{slade2022personalizing,molinaro2024task,liu2026exotraj}. Plantar-pressure sensors and surface electromyography (sEMG) provide additional information on foot contact and muscle activity, respectively.
Learning-based high-level controllers estimate locomotion states and biological joint moments or directly generate assistance commands~\cite{molinaro2024estimating}. These networks are typically trained offline and deployed onboard; some implementations use Open Neural Network Exchange (ONNX) export and TensorRT optimization~\cite{liu2026exotraj}. Jetson Nano and Orin NX modules support embedded inference, whereas Arduino and Teensy boards, STM32 microcontrollers, and platforms such as Raspberry Pi and NI myRIO support data acquisition and control.
Metabolic cost is commonly assessed through indirect calorimetry using systems such as the COSMED K5 and Parvo Medics TrueOne 2400~\cite{slade2022personalizing,lee2024universal}. Muscle activation is evaluated using sEMG systems, although hardware configurations vary across studies.
Onboard communication commonly relies on CAN for motor commands and state feedback. IMU data are acquired via wired interfaces such as SPI, I$^2$C, or UART, or via wireless links, while high- and low-level controllers exchange data through wired or serial connections.

\subsection{Adaptive Assistance}
Table~\ref{tab:multitask_exoskeletons} compares representative studies of adaptive lower-limb exoskeleton assistance in terms of hardware, hierarchical control, and experimental results. These implementations instantiate the four dimensions of embodied assistance as one pipeline, from structure and sensing to high-level decision-making and low-level actuation.
Hardware design, which underlies structural self-adaptation, must balance system mass, compliance, and real-time computational requirements~\cite{zhang2026enhancing}. Within this hardware, multimodal measurements are acquired according to the mechanical structure and sensor configuration~\cite{xu2026bioinspired}. Hierarchical control then separates the generation of assistance from its execution. The high-level controller uses AI algorithms to fuse these measurements into task-dependent assistance references (task adaptation)~\cite{molinaro2024task} and refines them according to user feedback (user-specific adaptation)~\cite{lee2023user}. The low-level controller tracks the resulting trajectories or torques with feedback- or optimization-based methods and maps them to motor outputs (control self-regulation)~\cite{liu2026exotraj}. Together, the four dimensions enable personalized and compliant assistance across complex tasks.

Assistance performance is evaluated by comparing the net metabolic rate, muscle activation, and heart rate under three conditions: walking without the exoskeleton (No Exo), wearing the exoskeleton with assistance disabled (Assist Off), and wearing it with assistance enabled (Assist On). Because these comparisons depend on the task and the participant population, Table~\ref{tab:multitask_exoskeletons} summarizes one of these outcomes—the change in net metabolic rate—for representative methods during level-ground walking, and reports the number of participants and the cohort type for each comparison. The increase in net metabolic rate in~\cite{molinaro2024task} is attributable to an exoskeleton mass of approximately 7~kg, which raises the wearer’s energy expenditure.
This mass penalty indicates what is still required for daily use: lightweight and biomechanically compatible exoskeletons that deliver reliable assistance across diverse activities and environments. A separate limitation concerns the measurement itself. Indirect calorimetry requires specialized equipment and prolonged measurement periods, so more practical assessment methods must be developed and validated.

    \section{Discussion and Conclusion}\label{sec_discus}
Embodied lower-limb exoskeletons form tightly coupled human--robot--environment systems. Their mechanical design should reconcile lightweight construction with biomechanical compatibility, and their assistance strategies should adapt to different environments, tasks, and users to provide effective and comfortable support. Existing studies have applied AI to predict motion trajectories or torques~\cite{molinaro2024task,luo2024experiment}, advancing assistive control in complex and dynamic environments. These strategies comprise two complementary layers. High-level representation integrates multimodal data to predict desired trajectories or torques, whereas low-level optimization refines assistive torques according to these predictions and predefined objectives.
The following discussion examines the limitations of existing work, and the corresponding directions for future research, along four dimensions.

\subsection{Existing Problems}
The control objectives of lower-limb exoskeletons differ for assistance in healthy adults and for rehabilitation in patients. This review focuses on assistance for healthy adults, where the objective is to augment lower-limb strength and reduce metabolic energy expenditure. Toward that objective, existing studies have advanced personalized assistance in complex environments. Along each of the four dimensions, that progress leaves a specific limitation.
\begin{enumerate}[leftmargin=*]
\item \textbf{Structural self-adaptation:} Mechanisms designed from joint anatomy improve kinematic compatibility~\cite{xu2026bioinspired}, and quasi-direct drive~\cite{huang2022modeling} and series elastic actuators~\cite{li2026spinal} are used to preserve motion compatibility. However, coordinated multi-joint assistance under a lightweight constraint remains difficult. Some of these designs increase exoskeleton mass, and the added mass can fatigue the wearer and reduce acceptance during prolonged use.

\item \textbf{Task adaptation:} Joint-torque data acquired with motion capture are fused to predict assistive torque~\cite{molinaro2024task,molinaro2024estimating}, and assistance policies trained in musculoskeletal simulation provide a second route to cross-task assistance~\cite{luo2024experiment}. However, motion-capture acquisition is costly and difficult to deploy on complex outdoor terrain, and musculoskeletal models for that terrain are difficult to train. These two constraints limit adaptation to complex tasks.

\item \textbf{User-specific adaptation:} Personalized parameters are optimized from physiological measurements~\cite{2022TROHILO,2025hip} or user feedback~\cite{lee2023user,slade2022personalizing}, and efficient optimization has been demonstrated with few interaction trials. However, extending these methods to high-dimensional parameters and to policy personalization for complex tasks, including neural-network policies, remains difficult.

\item \textbf{Control self-regulation:} Compensation-based~\cite{xing2024spatial} and optimization-based~\cite{li2022impedance} methods regulate motor outputs to track high-level commands. However, these methods rely on an accurate dynamic model. Because perception and control are tightly coupled, and the human–exoskeleton dynamics are difficult to model with high fidelity, tracking accuracy, robustness, and safety remain difficult to ensure.
\end{enumerate}

\subsection{Future Directions}
These challenges motivate further research in the following directions:
\begin{enumerate}[leftmargin=*]
\item \textbf{Mechanical design:} The mass of current multi-joint designs fatigues the wearer during prolonged use. Future exoskeletons should therefore be detachable, modular, and garment-like, with compliant attachment interfaces that reduce contact pressure and preserve biomechanical compatibility at low mass. Further mass reduction depends on lighter motors, batteries, and integrated circuits.

\item \textbf{Multimodal data acquisition:} Costly motion capture limits the data available for complex outdoor tasks. Lightweight sensing systems are needed so that these data can be collected in complex environments without specialized equipment. Mapping methods should then reconcile differences in sensor placement and specifications across exoskeletons, placing the heterogeneous recordings in one representation for training on complex terrain.

\item \textbf{Policy optimization frameworks:} High-dimensional policies, including neural-network policies, remain difficult to personalize for complex tasks. Preference-based frameworks should optimize them from real-time user feedback on assistance quality. In a tightly coupled human--robot system, this feedback keeps data collection short and limits user burden and fatigue.

\item \textbf{Human-robot dynamics modeling:} Coupled human--robot dynamics are difficult to capture analytically at the fidelity needed for accurate, robust, and safe tracking. Future research should learn data-driven models of these dynamics. The learned models could guide torque optimization and provide simulation environments for training assistance policies.
\end{enumerate}

These directions must be realized on a wearable device. Mechanical design and onboard algorithm deployment therefore need to achieve user acceptance, comfort during prolonged use, and portability. 
Broader sharing of datasets, hardware platforms, source code, and mechanical designs would allow these methods to be reproduced and extended, supporting more compliant human–robot interaction and more effective assistance. 
Beyond this exoskeleton setting, large multimodal models could use their language and visual understanding to carry the same framework into other human–robot interaction scenarios, where interaction can be more natural and efficient.

\subsection{Conclusion}

This review surveys lower-limb exoskeleton research from an embodied perspective and from the coupling among the human, the exoskeleton, and the environment. The literature is organized along four dimensions: structural self-adaptation, task adaptation, user-specific adaptation, and control self-regulation. Limitations remain in lightweight and compatible structural design, kinetic-data acquisition, personalized optimization of complex policies, and accurate modeling of coupled human–exoskeleton dynamics. Research on mechanical structure, perception algorithms, and control is still pursued largely independently. Future work should proceed from an embodied perspective and integrate these three aspects into an efficient and comfortable assistance solution for prolonged use. The review also provides a framework for analyzing related methods in human–robot interaction and physical assistance, including wearing comfort, task adaptability, user-specific adaptability, and assistance effectiveness.


\end{document}